\documentclass[twocolumn, switch]{article} 

\usepackage{comment}
\usepackage{svg}
\usepackage{bbding}
\usepackage{preprint}
\usepackage{pifont}
\usepackage{xcolor}
\usepackage{gensymb}

\usepackage{xspace}
\makeatletter
\DeclareRobustCommand\onedot{\futurelet\@let@token\@onedot}

\def\@onedot{\ifx\@let@token.\else.\null\fi\xspace}

\def\eg{\emph{e.g}\onedot} 
\def\ie{\emph{i.e}\onedot}

\def\etal{\emph{et al}\onedot}
\makeatother

\usepackage{amsmath, amsthm, amssymb, amsfonts}

\usepackage[numbers,square]{natbib}
\usepackage[utf8]{inputenc}	
\usepackage[T1]{fontenc}	
\usepackage{xcolor}		
\usepackage[colorlinks = true,
            linkcolor = purple,
            urlcolor  = blue,
            citecolor = cyan,
            anchorcolor = black]{hyperref}	
\usepackage{booktabs} 		
\usepackage{nicefrac}		
\usepackage{microtype}		
\usepackage{lineno}		
\usepackage{float}			

\usepackage{lipsum}		

\usepackage{newfloat}
\DeclareFloatingEnvironment[name={Supplementary Figure}]{suppfigure}
\usepackage{sidecap}
\sidecaptionvpos{figure}{c}

\usepackage{titlesec}
\titlespacing\section{0pt}{12pt plus 3pt minus 3pt}{1pt plus 1pt minus 1pt}
\titlespacing\subsection{0pt}{10pt plus 3pt minus 3pt}{1pt plus 1pt minus 1pt}
\titlespacing\subsubsection{0pt}{8pt plus 3pt minus 3pt}{1pt plus 1pt minus 1pt}

\usepackage{tikz,xcolor,hyperref}
\usepackage{cleveref}

\definecolor{lime}{HTML}{A6CE39}
\DeclareRobustCommand{\orcidicon}{
	\begin{tikzpicture}
	\draw[lime, fill=lime] (0,0) 
	circle [radius=0.16] 
	node[white] {{\fontfamily{qag}\selectfont \tiny ID}};
	\draw[white, fill=white] (-0.0625,0.095) 
	circle [radius=0.007];
	\end{tikzpicture}
	\hspace{-2mm}
}
\foreach \x in {A, ..., Z}{\expandafter\xdef\csname orcid\x\endcsname{\noexpand\href{https://orcid.org/\csname orcidauthor\x\endcsname}
			{\noexpand\orcidicon}}
}

\title{Leveraging Neural Radiance Fields for Pose Estimation of an Unknown Space Object during Proximity Operations}

\usepackage{authblk}

\author[1,2,4\thanks{\tt{antoine.legrand@uclouvain.be}}]{Antoine Legrand\orcidA{}}
\author[2,3]{Renaud Detry\orcidB{}}
\author[1]{Christophe De Vleeschouwer \orcidC{}}

\affil[1]{Department of Electrical Engineering (ELEN), ICTEAM, UCLouvain}
\affil[2]{Department of Electrical Engineering (ESAT), KU Leuven}
\affil[3]{Department of Mechanical Engineering (MECH), KU Leuven}
\affil[4]{Aerospacelab}

\begin{document}

\twocolumn[ 
  \begin{@twocolumnfalse} 
    \maketitle
    \begin{abstract}
    We address the estimation of the 6D pose of an unknown target spacecraft relative to a monocular camera, a key step towards the autonomous rendezvous and proximity operations required by future Active Debris Removal missions. We present a novel method that enables an "off-the-shelf" spacecraft pose estimator, which is supposed to known the target CAD model, to be applied on an unknown target. Our method relies on an in-the wild NeRF, \ie a Neural Radiance Field that employs learnable appearance embeddings to represent varying illumination conditions found in natural scenes. We train the NeRF model using a sparse collection of images that depict the target, and in turn generate a large dataset that is diverse both in terms of viewpoint and illumination. This dataset is then used to train the pose estimation network. We validate our method on the Hardware-In-the-Loop images of SPEED+~\cite{park2022speed+} that emulate lighting conditions close to those encountered on orbit. We demonstrate that our method successfully enables the training of an off-the-shelf spacecraft pose estimation network from a sparse set of images. Furthermore, we show that a network trained using our method performs similarly to a model trained on synthetic images generated using the CAD model of the target.
    \end{abstract}
  \end{@twocolumnfalse}] 

\section{Introduction}
\label{sec_intro}

    With an ever growing number of satellites in orbit, the risk of collision between a satellite and space debris, \eg, rocket bodies, defunct satellites or pieces from a previous collision, is steadily rising. Such a collision would not only cause the destruction of a functional satellite but also dramatically increase the number of space debris~\cite{rossi1998modelling}, thereby further increasing the risk of such a collision~\cite{kessler2010kessler}. As a result, private companies and space agencies are working on Active Debris Removal (ADR)~\cite{forshaw2020active,aglietti2020active,poozhiyil2023active} missions that aims at de-orbiting space debris. These ADR missions require to perform Rendezvous and Proximity Operations (RPO) with a non-cooperative target, \ie, a chaser spacecraft must operate close to, or even dock with, a target spacecraft which was not designed to support RPOs. 
    Due to the risk of human failures implied by tele-operated operations, those RPOs should be carried autonomously by the chaser spacecraft. 
    
\begin{figure}[t]
    \centering
    \includegraphics[width=0.99\linewidth]{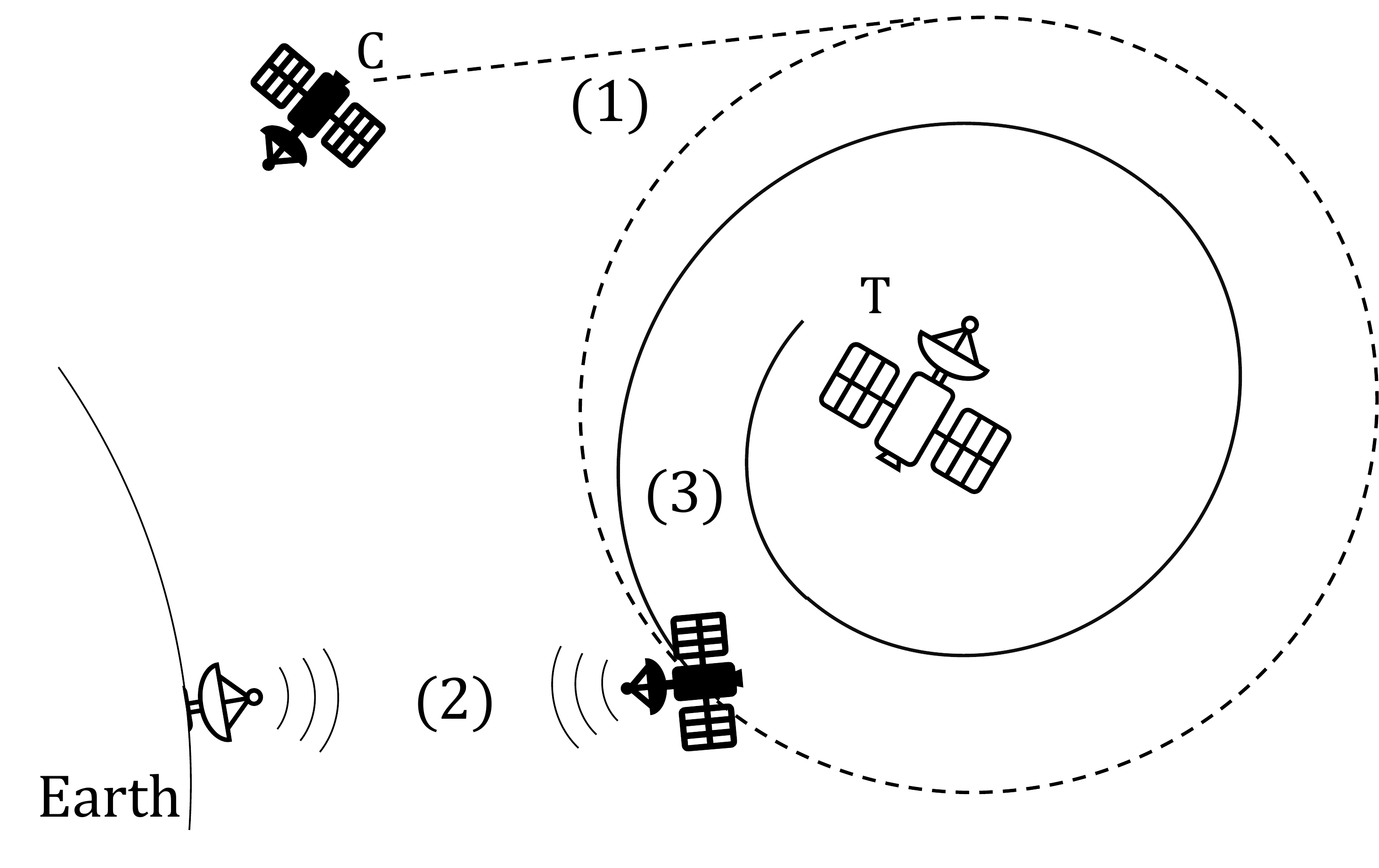}
    \caption{Overview of the considered three-steps operation. \textbf{(1)} The chaser spacecraft (C) approaches the target (T) while taking pictures. \textbf{(2)} The images are downloaded on ground and processed to train a Spacecraft Pose Estimation network, whose weights are uploaded on the chaser. \textbf{(3)} The chaser finishes the operation autonomously, by relying on the trained network.}
    \label{fig_overview_semiauto_rpo}
    \vspace{-0.5cm}
\end{figure}

    A key capability to perform autonomous RPOs is the on-board estimation of the relative pose, \ie, position and orientation of the target spacecraft relative to the chaser. Due to their low cost, low mass and compactness, monocular cameras are considered for this task~\cite{d2014pose,pauly2023survey}. Although the vision-based estimation of the relative pose of a non-cooperative spacecraft has been studied in depth in the literature~\cite{park2023robust,chen2019satellite,proencca2020deep}, current solutions assume the knowledge of the CAD model of the target spacecraft that enables the generation of large synthetic training sets. In the case of Active Debris Removal, this assumption does not hold since little information is known about the debris. This work aims at leveraging Neural Radiance Fields (NeRFs) models~\cite{mildenhall2021nerf} to extend the scope of existing pose estimation methods to unknown targets, \ie, targets for which the CAD model is not available. 

    For this purpose, we consider a three-steps approach, as illustrated in \Cref{fig_overview_semiauto_rpo}. Firstly, the chaser spacecraft is tele-operated to approach the target up to a safe distance. During the approach, the chaser acquires images of the target and transmit them to a ground station. Then, those few images are processed on-ground to synthesize additional views of the target under varying illuminations, so as to build a sufficiently rich set of images to train an "off-the-shelf" pose estimation network, \ie an existing neural network that only requires to be trained on a new set depicting the target. Finally, the model weights are uploaded on the spacecraft which autonomously performs the final approach. The on-ground processing step enables the use of the virtually infinite computing resources available on ground, in contrast with the low-power on-board hardware. Furthermore, even if the chaser spacecraft requires ground support in this scenario, it operates autonomously during the critical, \ie, close-range, phase of the operation. 

    To demonstrate the feasibility of such a 3-steps procedure in terms of image analysis requirements, this paper aims at studying  the performance obtained when training a pose estimator from a few images. Hence, it focuses on the on-ground processing step, \ie on the training of a spacecraft pose estimation model from a small number of spaceborne images depicting that target spacecraft. For this purpose, our method resorts to an implicit representation of the target, under the form of an "in-the-wild" Neural Radiance Field~\cite{mildenhall2021nerf,fridovich2023k,martin2021nerf}. This NeRF is then used to generate a sufficiently large training set which captures the diversity of both the pose distribution and the illumination conditions encountered in orbit. Finally, an off-the-shelf Spacecraft Pose Estimation (SPE) network is trained on this set. As pointed out in \Cref{tab_position_sota}, our work is the first to validate a vision-based spacecraft pose estimation method for unknown targets on realistic images, \ie, the Hardware-in-the-loop images of SPEED+~\cite{park2022speed+}.  

\begin{table}[t]
    \centering
    \begin{tabular}{c|cc}
        \toprule
        SoA SPE methods & Model- & Validation on  \\
         & agnostic & realistic images \\   
        \midrule
        known target~\cite{park2023robust,chen2019satellite,proencca2020deep} &  & \ding{51} \\
        unknown target~\cite{huang2021low,ren2020pose,park2024rapid} & \ding{51} &  \\
        \textbf{Ours} & \ding{51} & \ding{51}\\
        \bottomrule
    \end{tabular}
    \vspace{0.3cm}
    \caption{Overview of the State-of-The-Art in the field of Spacecraft Pose Estimation. Our method is the first model-agnostic one to be validated on realistic images.}
    \label{tab_position_sota}
    \vspace{-0.4cm}
\end{table}

\section{Spacecraft Pose Estimation}
    
    The existing methods for spacecraft pose estimation are either model-agnostic or model-based. While the first ones do not assume any prior on the target spacecraft, the latter ones exploit the CAD model of the target. 

    Regarding the model-agnostic methods, most works rely on (i) sensors such as stereo~\cite{pesce2017stereovision,feng2018pose} or Time-of-Flight~\cite{guo2020real} cameras or (ii) on multiple chaser spacecraft~\cite{matsuka2021collaborative}. A few works~\cite{huang2021low,ren2020pose} tackled the problem using a monocular camera while relying on keypoint detection and description~\cite{huang2021low,harris1988combined,lowe2004distinctive,rublee2011orb} or edge detection~\cite{ren2020pose} to estimate the pose from the observed image. Apart from those keypoint/edge detection methods, Park \etal~\cite{park2024rapid} proposed a CNN-based method to recover both the shape and the pose of an unknown spacecraft from a single image. However, those works were only validated on synthetic images. Unlike them, our method, which extends the applicability of model-based methods to an unknown target observed from a sparse set of viewpoints, is successfully demonstrated on Hardware-In-the-Loop (HIL) images from the SPEED+ dataset~\cite{park2022speed+}.      
    

    Model-based methods received a strong interest in recent years, notably through the Spacecraft Pose Estimation Challenges (SPECs)~\cite{kisantal2020satellite,park2023satellite}. State-of-the-Art solutions rely on Convolutional Neural Networks (CNN) to either directly estimate the spacecraft pose from the image~\cite{sharma2018pose,proencca2020deep,park2023robust} or predict keypoint coordinates~\cite{chen2019satellite,park2019towards,legrand2022end} which are then used to recover the pose by solving the Perspective-n-Points (PnP) problem~\cite{lepetit2009ep}. While most works~\cite{chen2019satellite,park2019towards,legrand2022end} rely on an intermediate step of spacecraft detection to identify a Region-of-Interest to crop and process through the CNN, some works~\cite{proencca2020deep,park2023robust} directly perform the estimation on the whole image. In all cases, those methods always rely on a large synthetic training set generated using the CAD model of the target spacecraft and therefore implicitly assume the knowledge of this CAD model, making them unsuitable in front of unknown targets. 
    
    Our work leverages a Neural radiance Field~\cite{mildenhall2021nerf} (NeRF) trained on a few spaceborne images to generate the large training set required by model-based Spacecraft Pose Estimation (SPE) pipelines. Thereby, it makes an arbitrary off-the-shelf, model-based, SPE network relevant to infer the pose of an unknown target, only characterized by a few image samples.

\begin{figure}[t]
    \centering
    \includegraphics[width=0.925\linewidth]{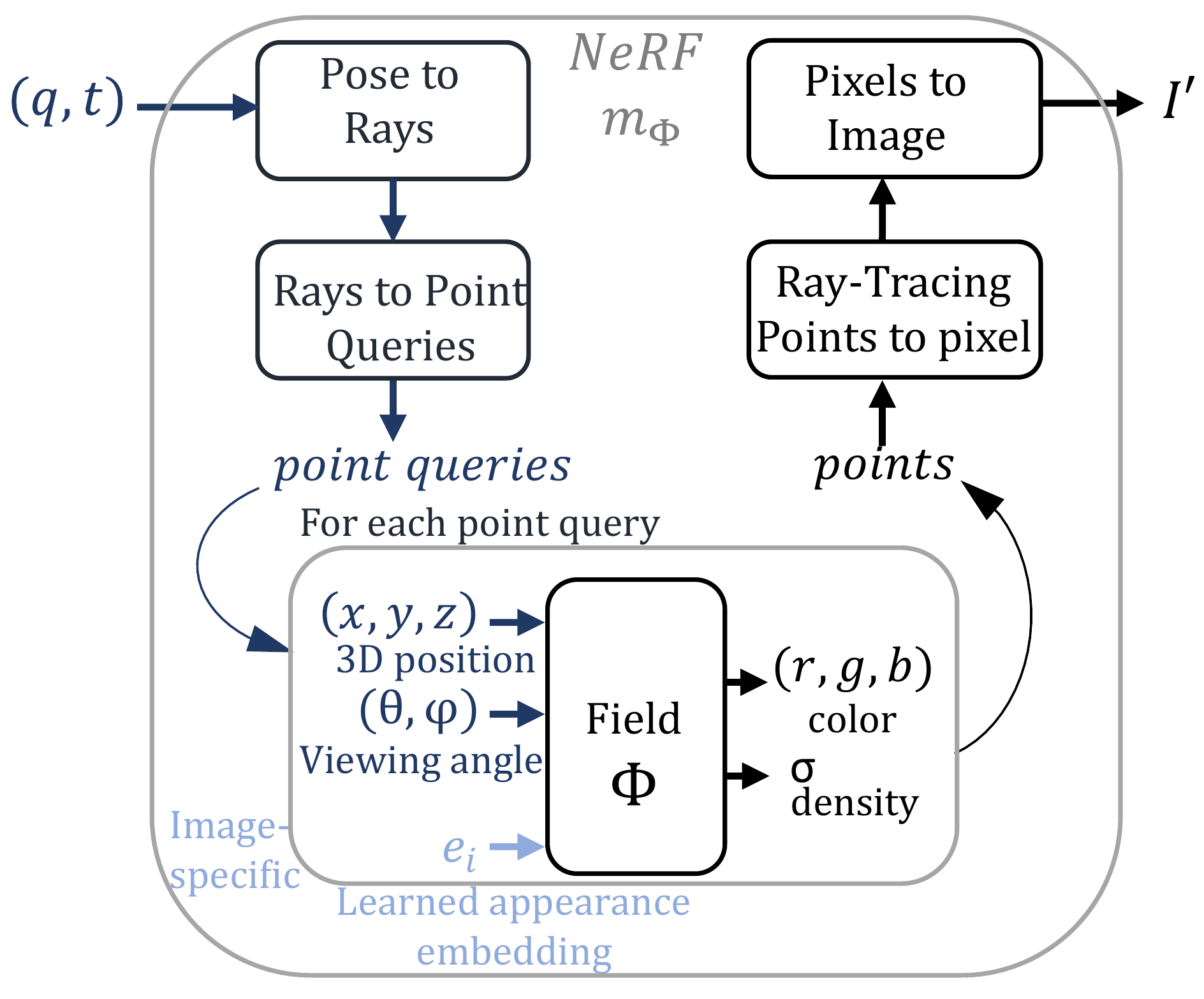}
    \caption{Overview of the generation of an image $I'= m_{\Phi}(q,t)$ through an in-the-wild Neural Radiance Field~\cite{martin2021nerf,fridovich2023k} $m_{\Phi}$ for a given pose $(q,t)$. As explained in \cref{sec_nerf_background}, given a pose, \ie rotation $q$ and translation $t$, the NeRF renders an image $I'$ by querying multiple times a MLP, which approximates the radiance field, and by aggregating the predicted color and density of the points through ray-tracing techniques. Unlike most NeRFs where the MLP takes as input a 3D position and 2 viewing angles to output the color and density of each point, In-the-wild NeRFs~\cite{martin2021nerf} also feeds the MLP with a learnable appearance embedding which is specific to each image. This offers the possibility to change the illumination conditions when synthesizing new images (see \cref{sec_appearance_embeddings}).}
    \label{fig_in_the_wild_nerf}
\end{figure}

\section{Neural Radiance Fields}
\label{sec_nerf_background}
    Neural Radiance Fields~\cite{mildenhall2021nerf} (NeRF) were developed as a tool to render novel views of a scene. In a NeRF, the scene is implicitly represented by a learned neural network. To render a novel view, the image is processed pixel per pixel. For each pixel, a ray is projected in the scene and several points are sampled along that ray. Each point is determined by a 3D position and 2 viewing angles which are fed in a MLP that outputs an rgb triplet and a volume density $\sigma$. The value of the pixel is recovered through differentiable ray-tracing techniques that aggregate all the points along the ray. To train this implicit representation, a reconstruction loss, computed between the rendered image and the corresponding ground-truth, is back-propagated through the neural network. 

    Improvements to the original NeRF method~\cite{mildenhall2021nerf} have been proposed regarding the training time~\cite{muller2022instant} or the ability to represent real scenes~\cite{martin2021nerf}. NeRF in the Wild~\cite{martin2021nerf} has introduced learnable appearance embeddings that are associated to each training image. As depicted in \Cref{fig_in_the_wild_nerf}, the embedding is given to the MLP along with the input coordinates, thereby allowing the NeRF to render the same view under different appearances trough the appearance embeddings. In this work, as both the fast training and the ability to represent a real scene are necessary, $K$-Planes~\cite{fridovich2023k} is used as it combines an efficient encoding with learnable appearance embeddings. 

    Neural Radiance Fields are used in many fields, \eg robotics~\cite{adamkiewicz2022vision,rosinol2023nerf}, virtual reality~\cite{deng2022fov}, and used for diverse  tasks such as novel views rendering~\cite{mildenhall2021nerf}, visual scene understanding~\cite{nguyen2024semantically} or pose estimation~\cite{yen2021inerf}. In the aerospace field, NeRFs have been used to render novel views of Mars~\cite{giusti2022marf}, recover the 3D shape of space objects~\cite{mergy2021vision,caruso20233d} or estimate the state of a spacecraft~\cite{heintz2023spacecraft}. However, they have never been used as a tool enabling the training of a Spacecraft Pose Estimation network on an unknown target.

\section{Method}
    As stated in the introduction, this paper provides a method that enables the use of an off-the-shelf, model-based, Spacecraft Pose Estimation (SPE) network on an unknown target in the context of a semi-autonomous Rendezvous and Proximity Operation (RPO).
\begin{figure}[t]
    \centering
    \includegraphics[width=0.9\linewidth]{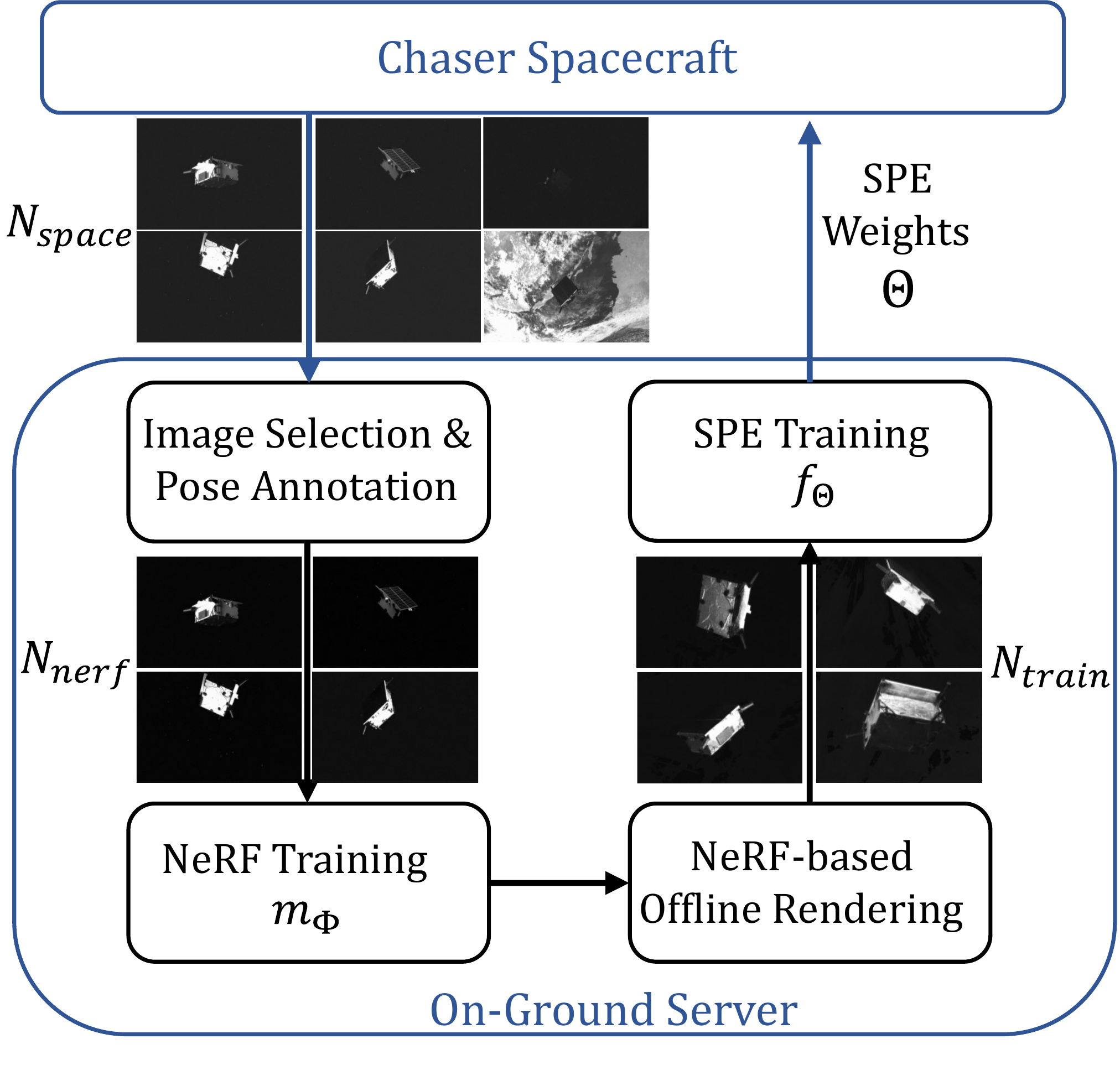}
    \caption{On-ground processing pipeline. A subset of the images downloaded from the spacecraft is annotated and used to train a NeRF~\cite{mildenhall2021nerf}, $m_{\Phi}$. This radiance field is then used to render a large training set, which is exploited to train an off-the-shelf SPE network $f_{\Theta}$. Finally, the network weights, $\Theta$, are uploaded on the spacecraft. Variables are defined in the text.}
    \label{fig_on_ground_processing}
\end{figure}
    
\subsection{Overview}
     As illustrated in \Cref{fig_overview_semiauto_rpo}, the considered RPO is made of 3 steps. First, the chaser spacecraft is tele-operated to approach the target and take pictures that are transferred to a ground station. On ground, the images are processed to train a SPE network whose weights are then uploaded on the chaser spacecraft. Finally, the chaser performs the final approach autonomously by exploiting the so trained pose estimation network. 
    
    This section describes the on-ground processing required to train an off-the-shelf spacecraft pose estimation model from a sparse set of spaceborne images. As depicted in \Cref{fig_on_ground_processing}, $N_{space}$ images are downloaded from the chaser spacecraft. From this set, $N_{nerf}$ high-quality images, \ie with good illumination conditions, are selected and their pose is annotated. They are then used to train a Neural Radiance Field (NeRF)~\cite{mildenhall2021nerf} $m_{\Phi}$ that learns an implicit representation of the target spacecraft. This radiance field is then used to generate a training set made of $N_{train}$ images which is used to train an off-the-shelf SPE network $f_{\Theta}$ whose weights $\Theta$ are finally uploaded on the chaser spacecraft. Those steps are detailed in the following sections.

\subsection{Images Selection and Pose Annotation}
\label{sec_method_selection}

    Due to the harsh lighting conditions encountered in orbit, some of the downloaded images can be over-exposed or under-exposed. As those images contain little information and would act as a noisy and misleading supervision in the NeRF training, they are discarded. Similarly, all the images where the Earth appears in the background are removed. Indeed, in a field aligned with the target, the Earth is a transient object which can not be explained by the NeRF. Since exploiting those images to train the NeRF would introduce significant artifacts, they are simply discarded. Finally, each image is annotated with pose information. 

\subsection{NeRF Training}
    Using 90\% of the $N_{nerf}$ images, we train an "in-the-wild" NeRF~\cite{martin2021nerf,fridovich2023k} $m_{\Phi}$, \ie, a Neural Radiance Field that contains learnable appearance embeddings as illustrated in \Cref{fig_in_the_wild_nerf}. Those embeddings enable the network to capture illumination conditions that are specific to each image and therefore render images with a larger illumination diversity.

\begin{figure}[t]  
    \includegraphics[width=.2475\linewidth]{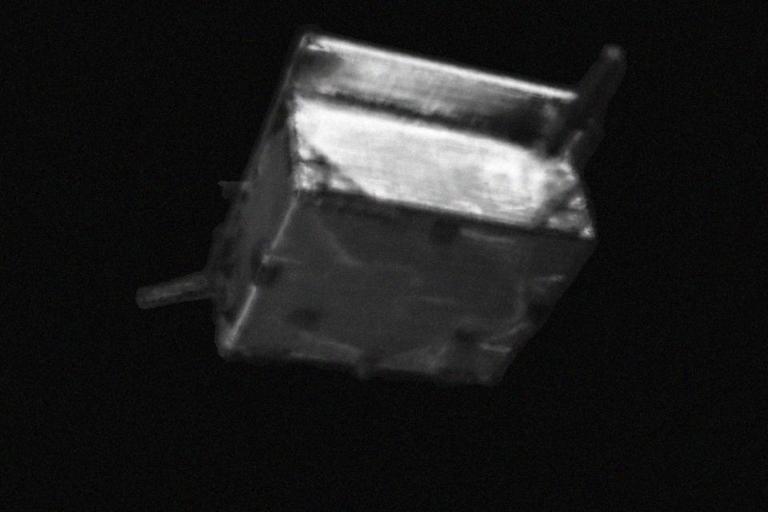}\hfill
    \includegraphics[width=.2475\linewidth]{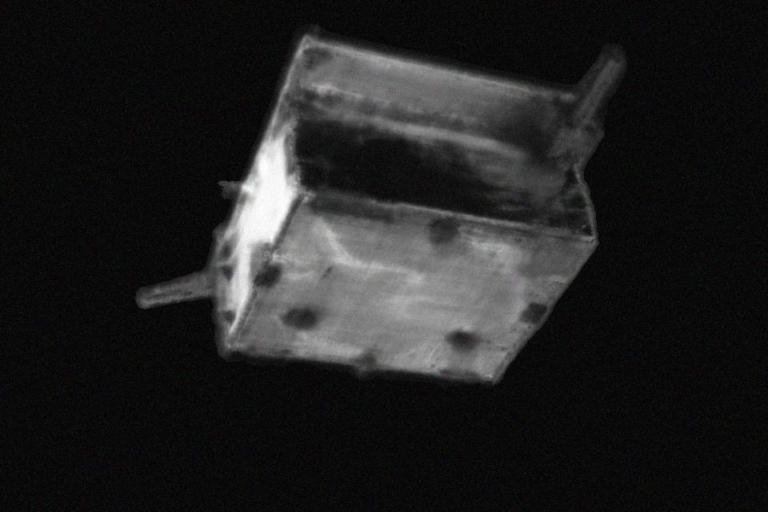}\hfill
    \includegraphics[width=.2475\linewidth]{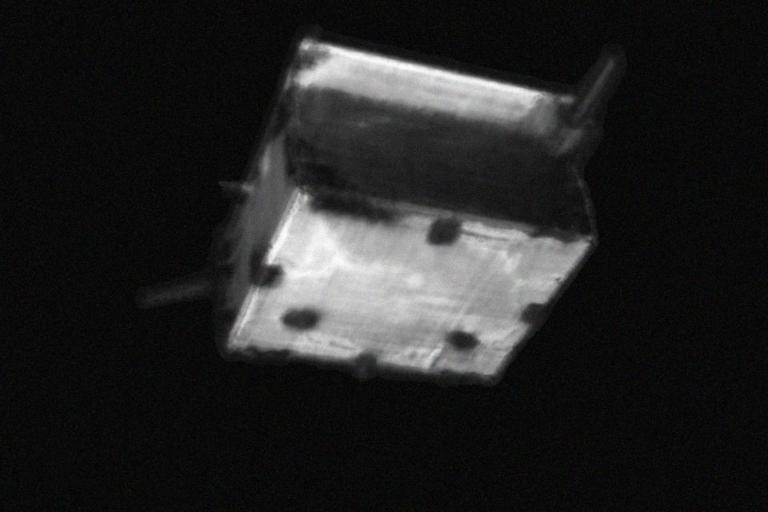}\hfill
    \includegraphics[width=.2475\linewidth]{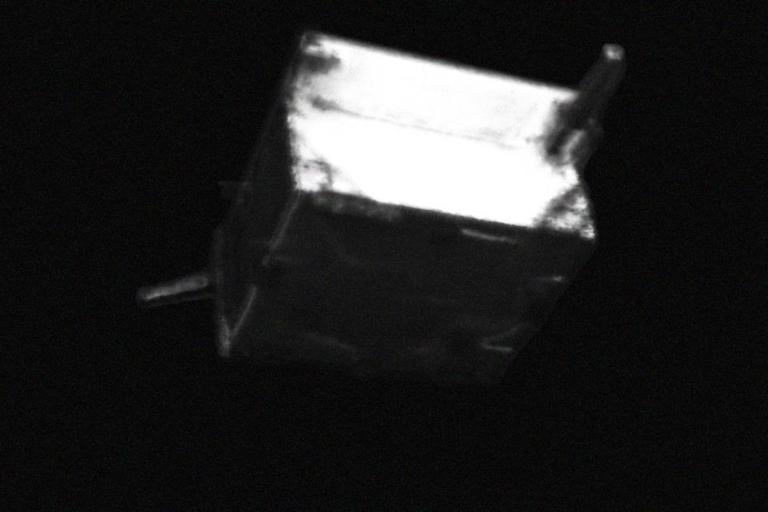}\\    
    \includegraphics[width=.2475\linewidth]{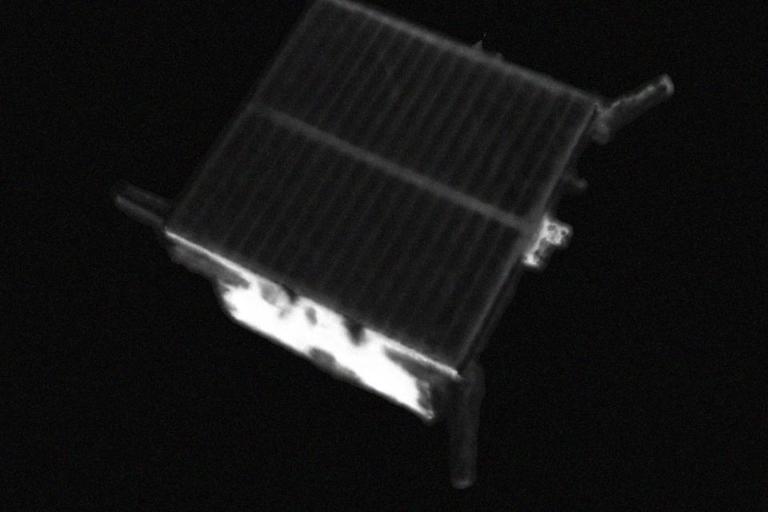}\hfill
    \includegraphics[width=.2475\linewidth]{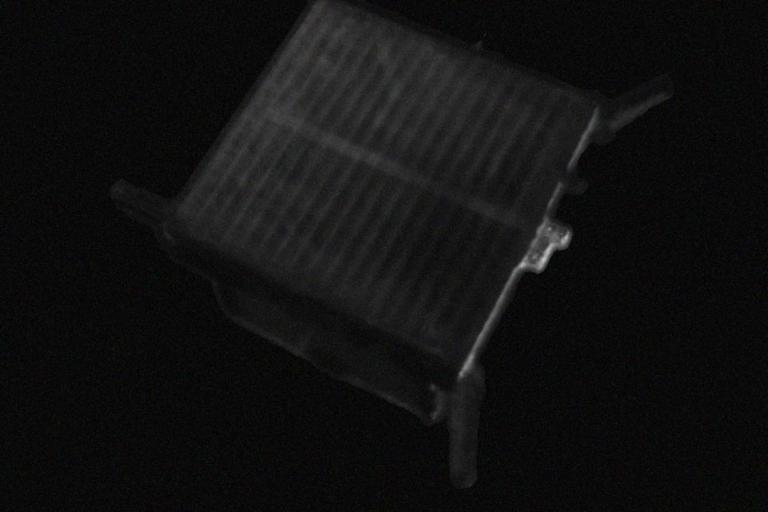}\hfill
    \includegraphics[width=.2475\linewidth]{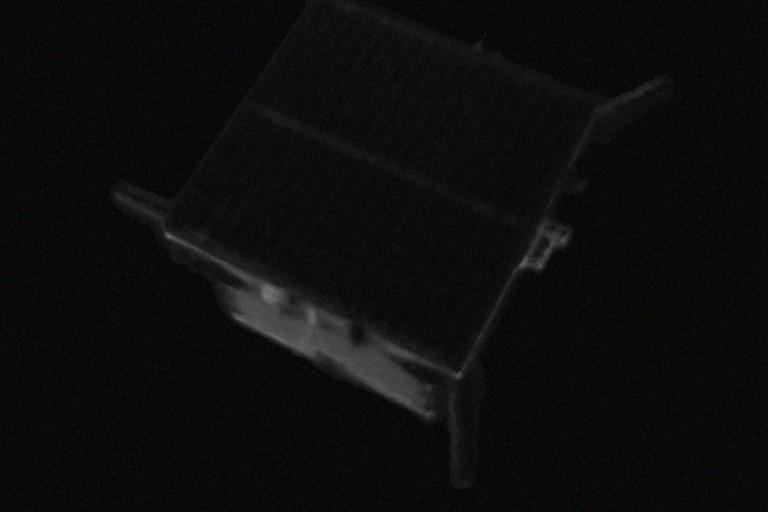}\hfill
    \includegraphics[width=.2475\linewidth]{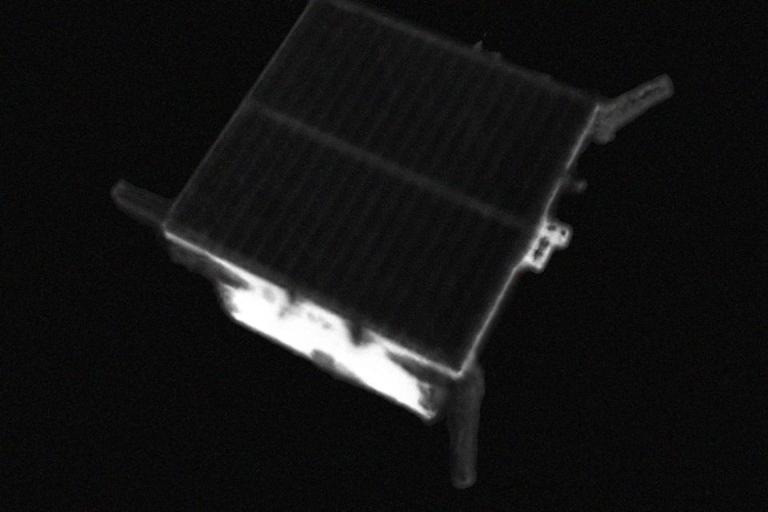}\\    
    \includegraphics[width=.2475\linewidth]{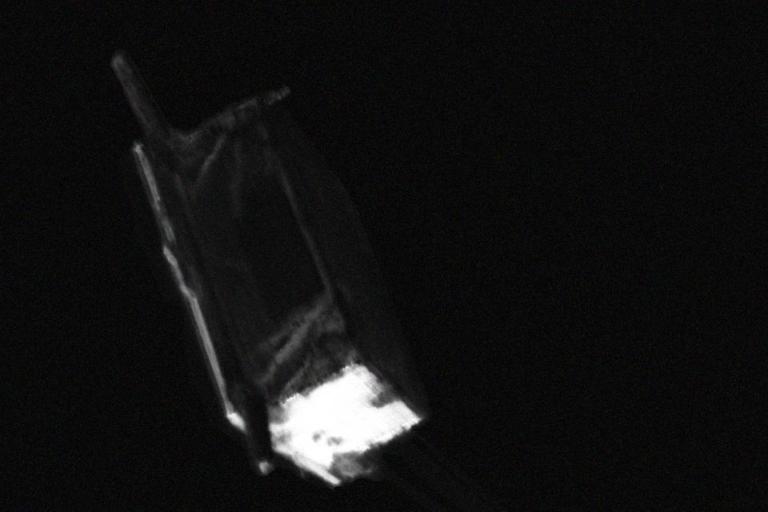}\hfill
    \includegraphics[width=.2475\linewidth]{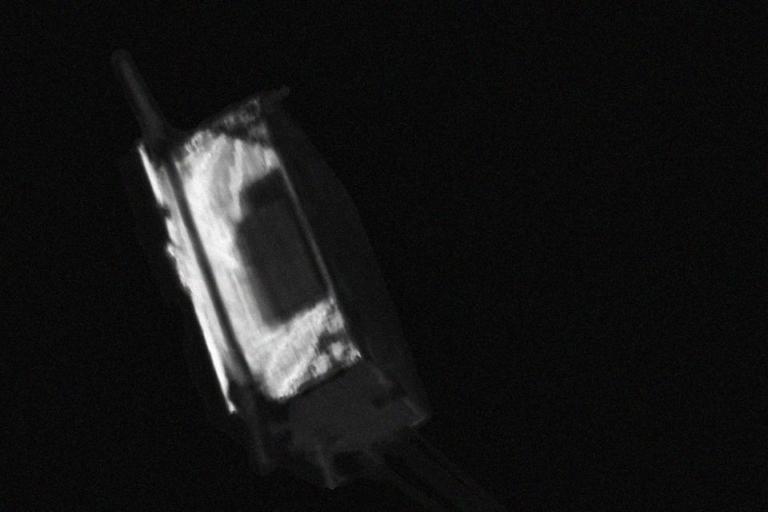}\hfill
    \includegraphics[width=.2475\linewidth]{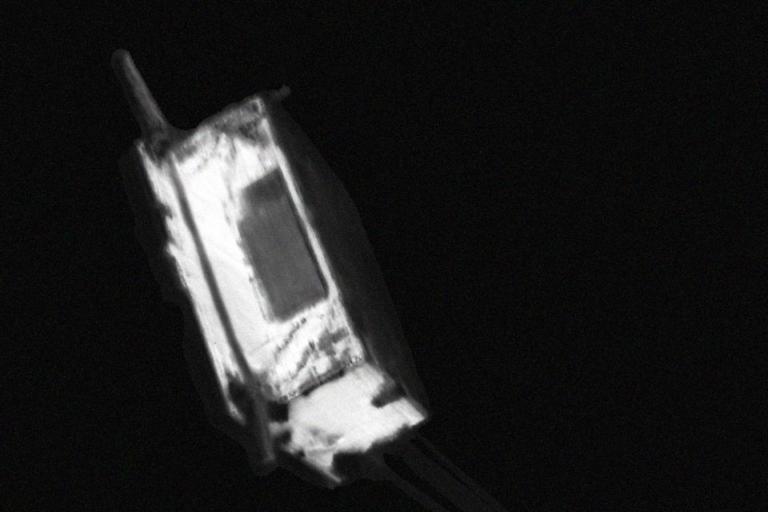}\hfill
    \includegraphics[width=.2475\linewidth]{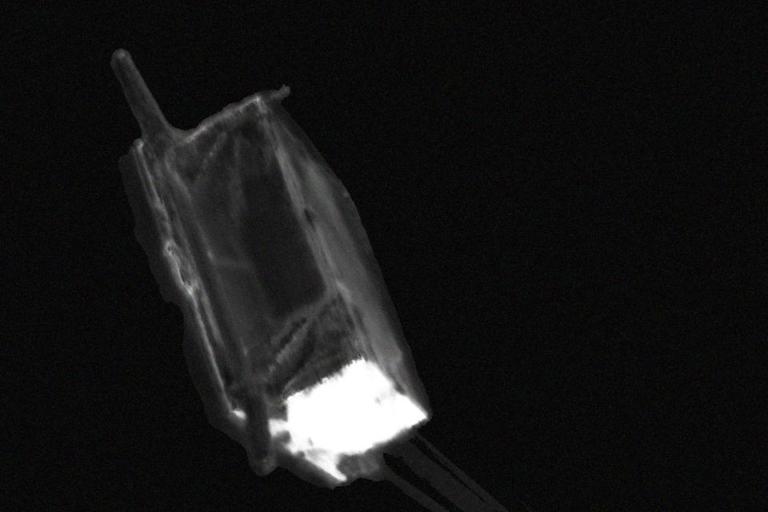}\\
    \caption{Images rendered by the same NeRF using 4 different appearance embeddings (columns 1-4). Randomizing the appearance embeddings enables the generation of images with diverse illumination conditions, regarding both the intensity and the orientation of the illumination source.}
    \label{fig_nerf_appearance_diversity}
\end{figure}

 \begin{table*}[t]
    \centering
    \begin{tabular}{c|c|c|c|c|c}
        \toprule
        & \textit{Sunlamp}~\cite{park2022speed+} & \multicolumn{2}{c|}{\textit{Lightbox}~\cite{park2022speed+}} & our & synthetic\\
        \cline{3-4}
        & & \textit{Lightbox*} & \textit{Lightbox-500} & training set & training set~\cite{park2022speed+}\\
        \midrule
        Usage & SPE test & SPE test & NeRF-training & SPE train & SPE train\\
        Domain & HIL  & HIL  & HIL   & NeRF-based & Synthetic\\
        Illumination& Direct & Diffuse & Diffuse & Randomized & Synthetic \\
        \# images & 2791 & 6240 & 500 & 48,000 & 48,000\\
        Examples & \includegraphics[width=.15\linewidth]{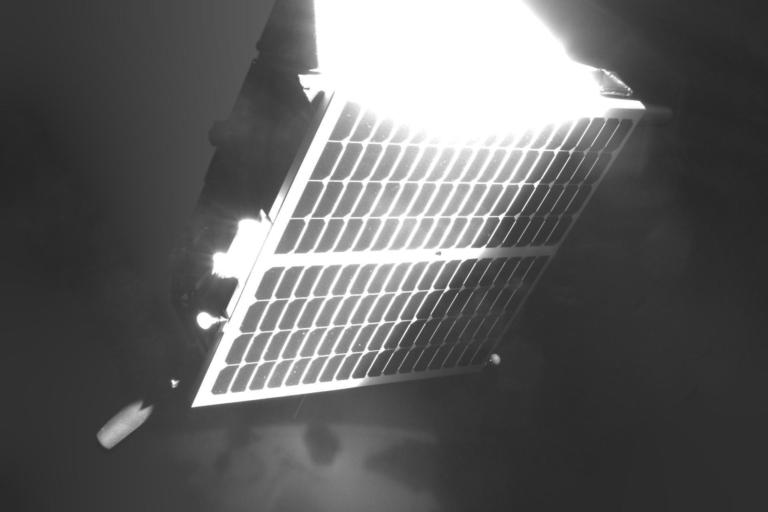} & \includegraphics[width=.15\linewidth]{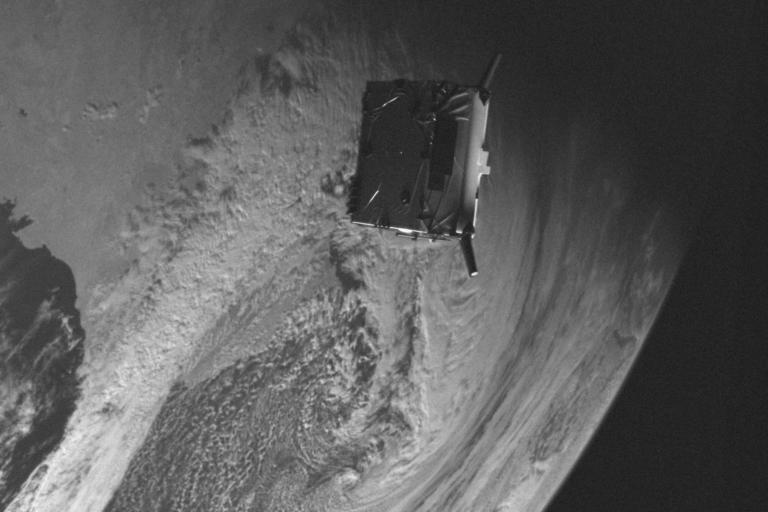} & \includegraphics[width=.15\linewidth]{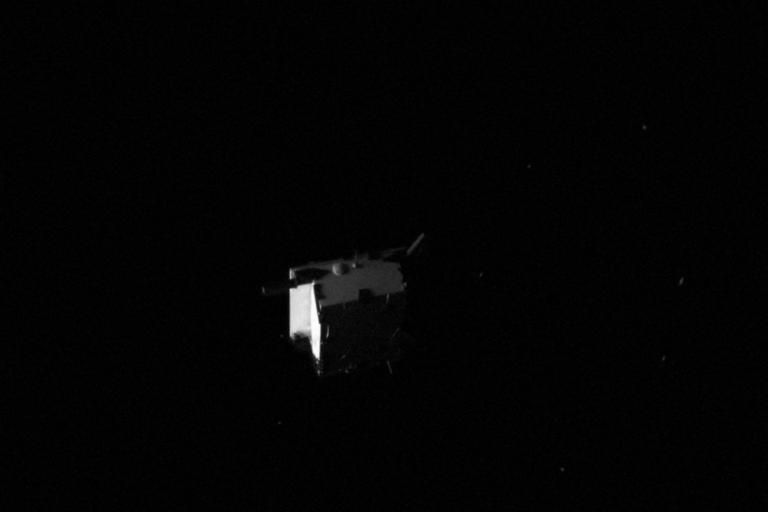} & \includegraphics[width=.15\linewidth]{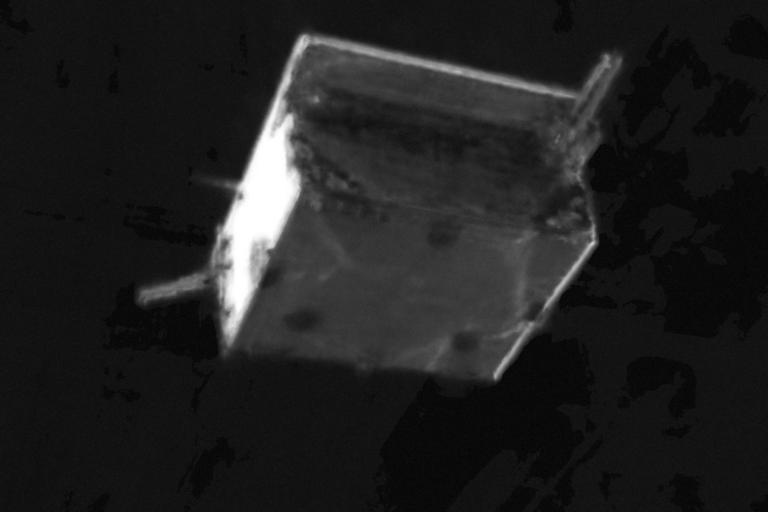} & \includegraphics[width=.15\linewidth]{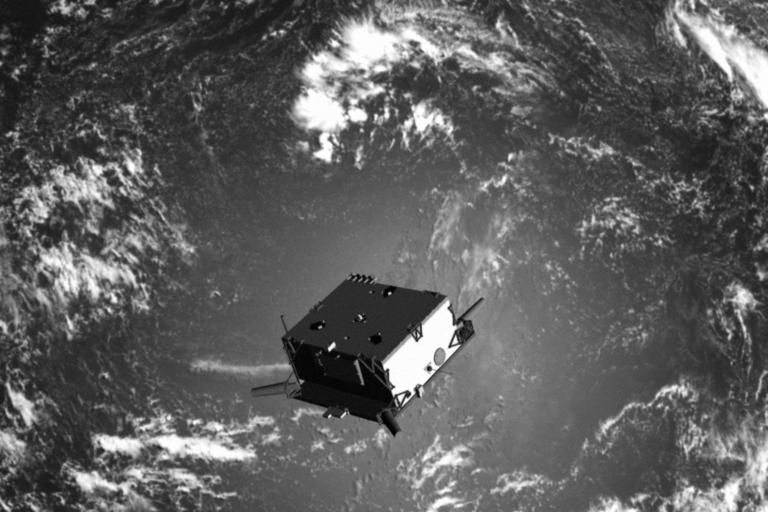}\\
         & \includegraphics[width=.15\linewidth]{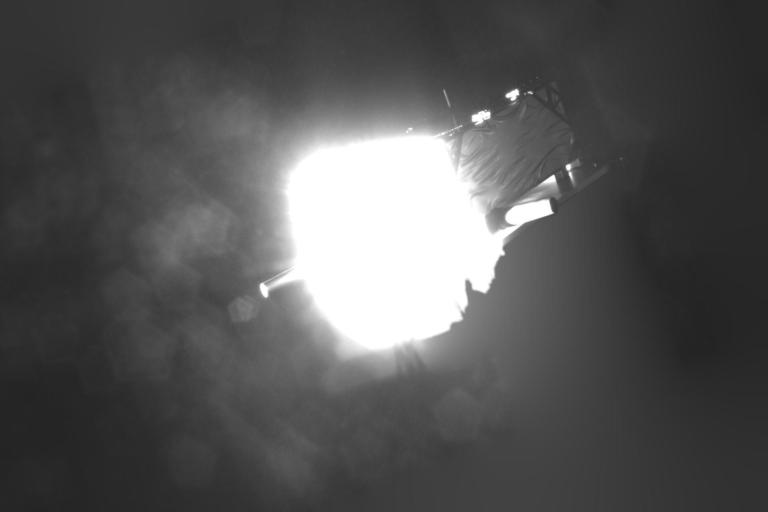} & \includegraphics[width=.15\linewidth]{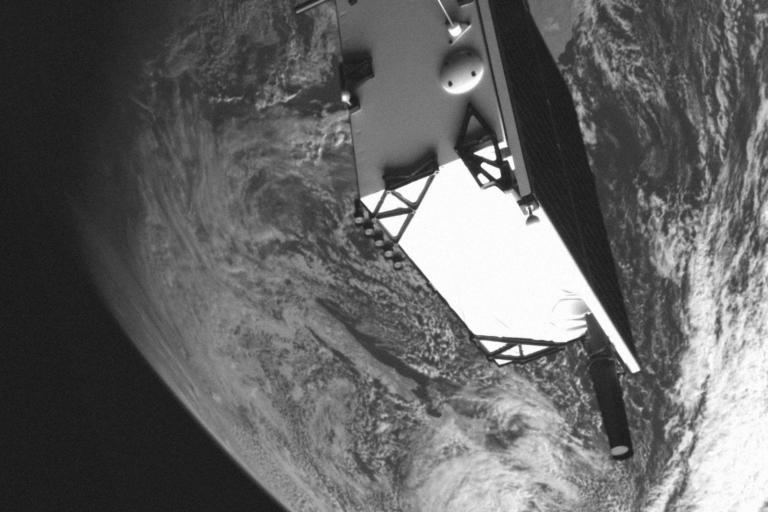}  & \includegraphics[width=.15\linewidth]{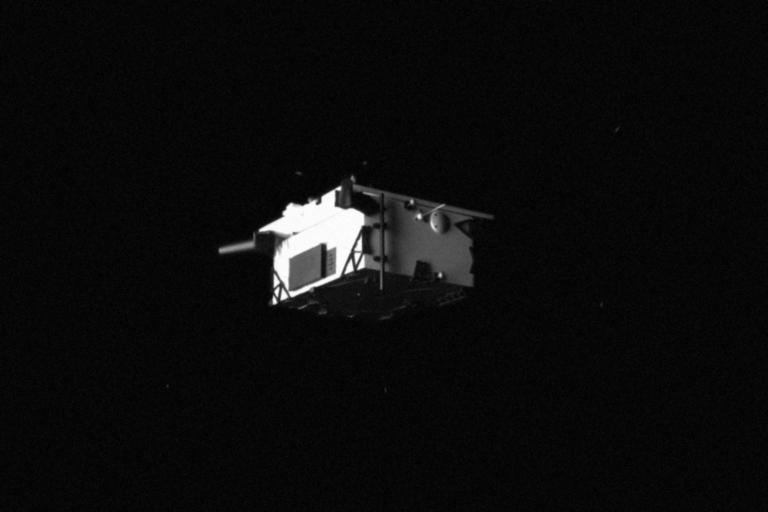}  & \includegraphics[width=.15\linewidth]{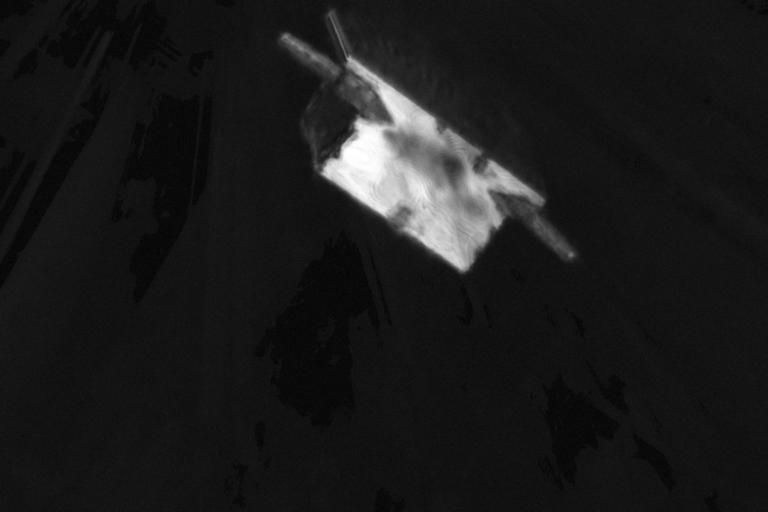} & \includegraphics[width=.15\linewidth]{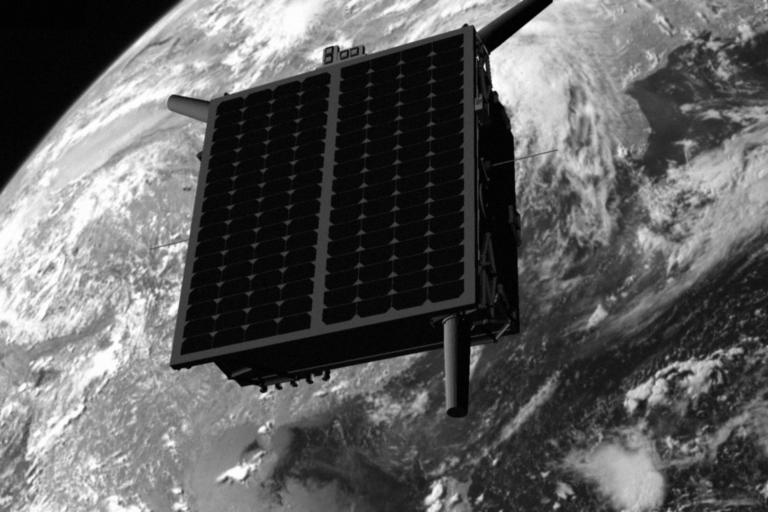}\\
         & \includegraphics[width=.15\linewidth]{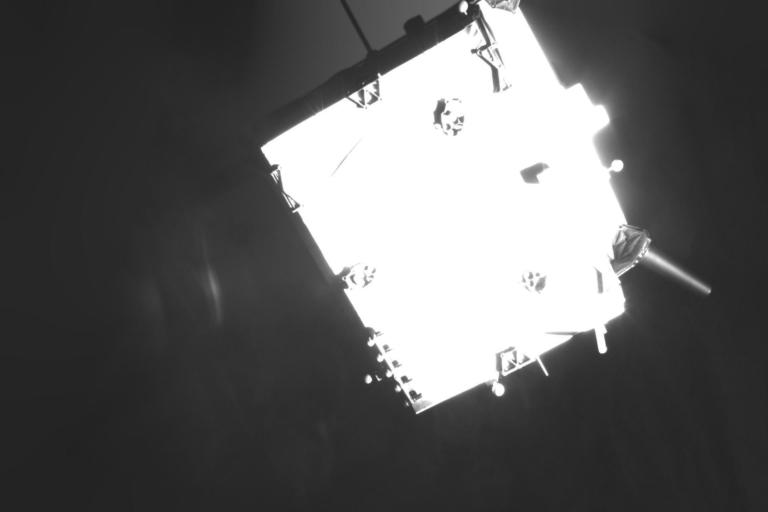} & \includegraphics[width=.15\linewidth]{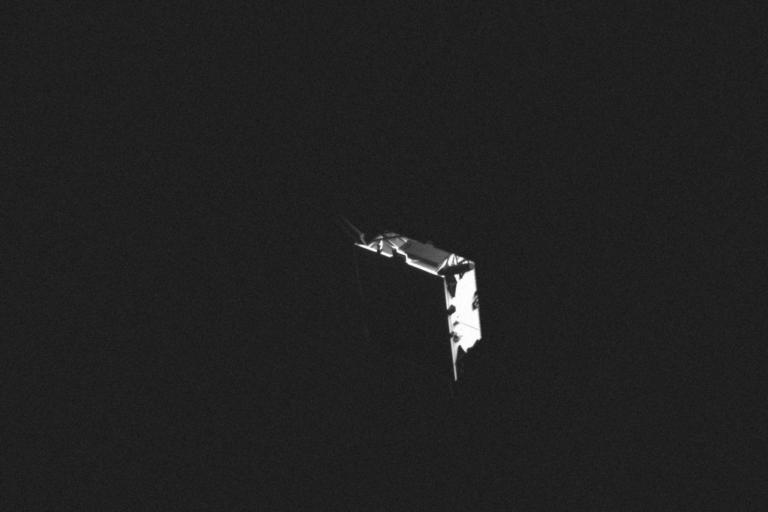}  & \includegraphics[width=.15\linewidth]{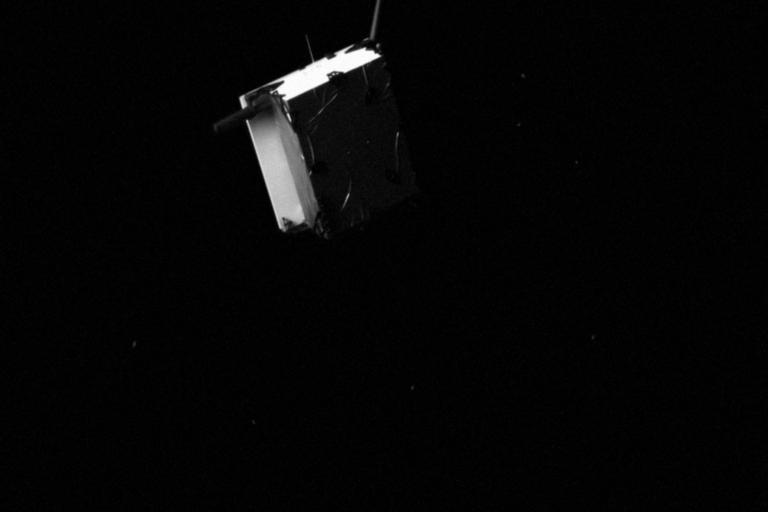}  & \includegraphics[width=.15\linewidth]{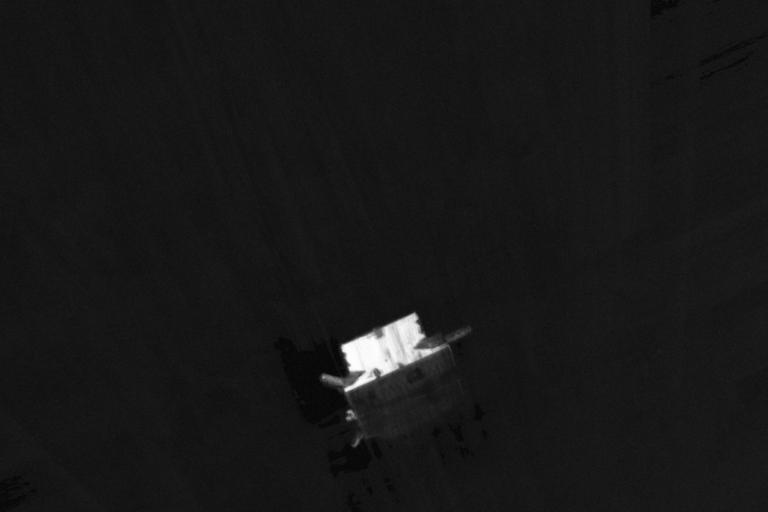} & \includegraphics[width=.15\linewidth]{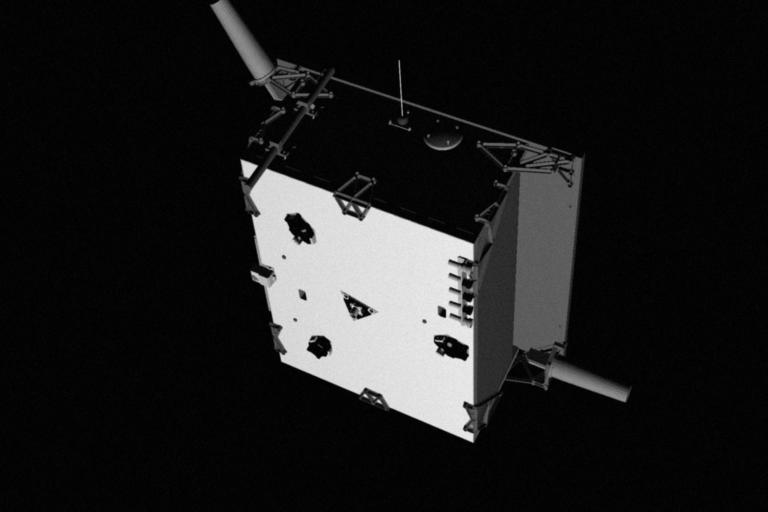}\\
         & \includegraphics[width=.15\linewidth]{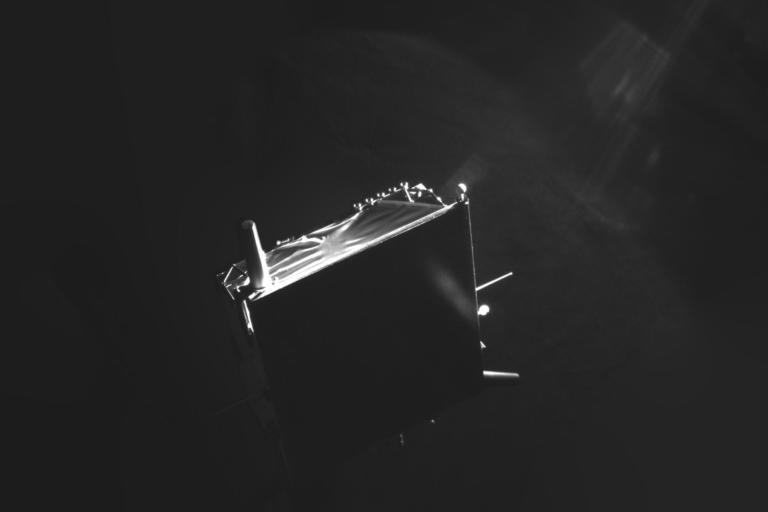} & \includegraphics[width=.15\linewidth]{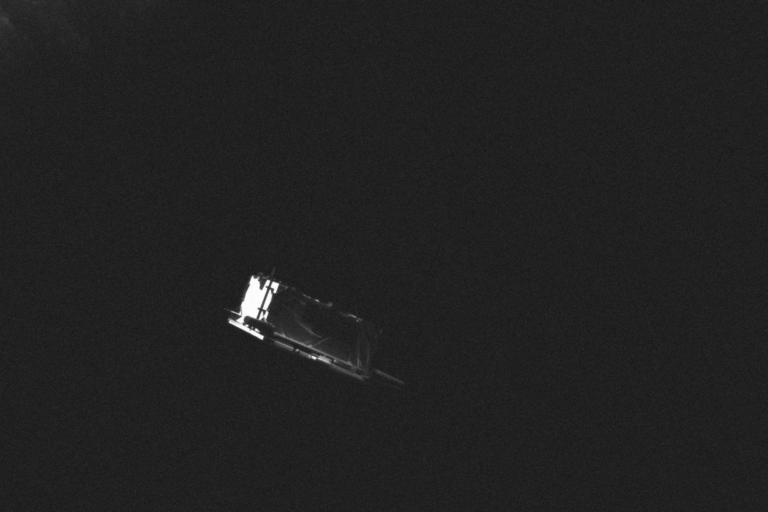}  & \includegraphics[width=.15\linewidth]{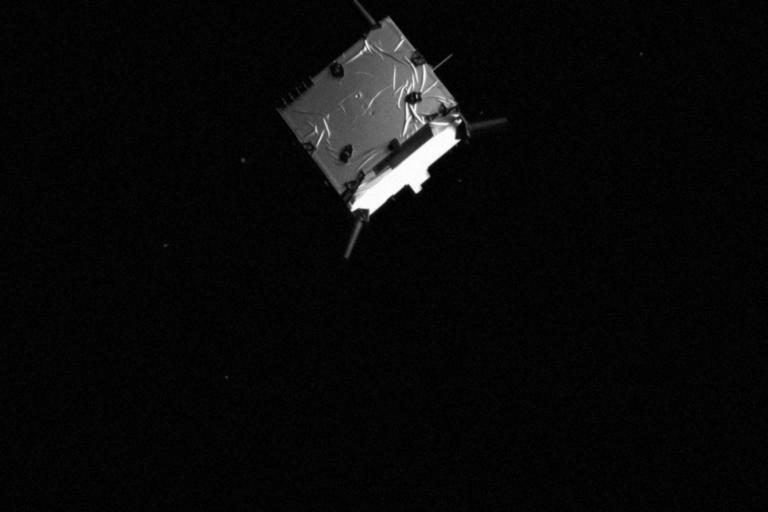}  & \includegraphics[width=.15\linewidth]{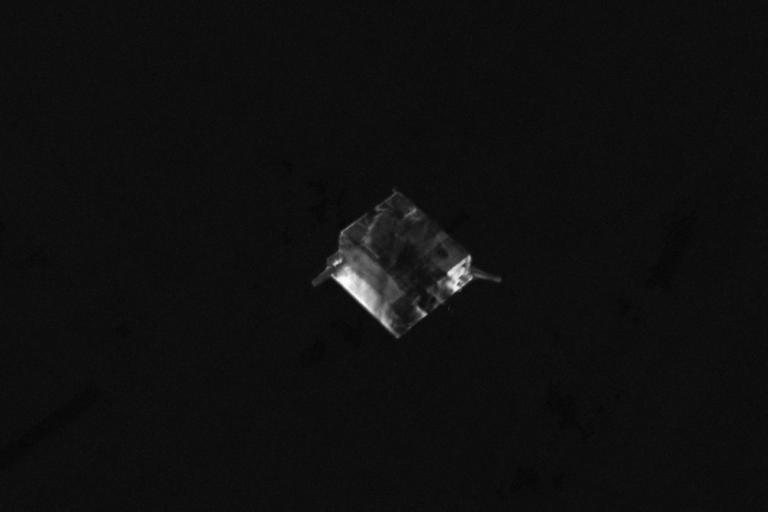} & \includegraphics[width=.15\linewidth]{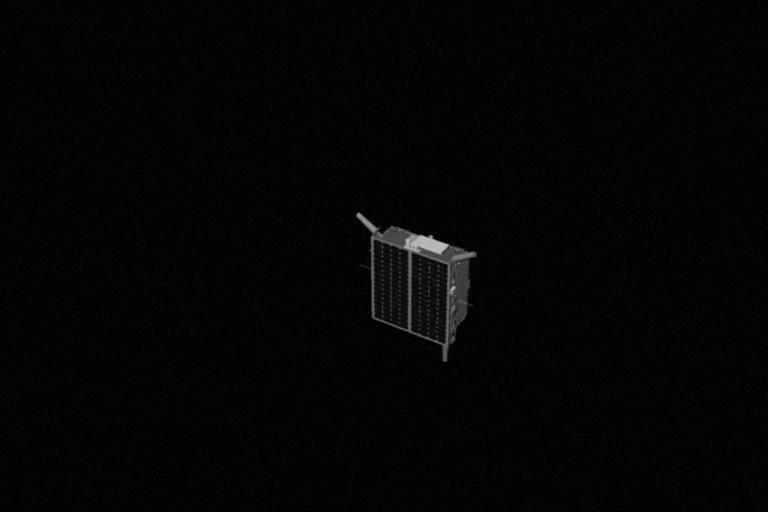} \\            
        \bottomrule
    \end{tabular}
    \caption{Overview of the sets used in this paper. \textit{Sunlamp} and \textit{Lightbox} are two Hardware-In-the-Loop (HIL) sets~\cite{park2022speed+}. \textit{Lightbox} is divided into \textit{Lightbox*} and \textit{Lightbox-500}. \textit{Lightbox-500} contains 500 images, selected as explained in \cref{sec_method_selection}, while \textit{Lightbox*} contains all the remaining ones. \textit{Sunlamp} and \textit{Lightbox*} are only used for testing. Because of the direct illumination of the Sun, which causes over-exposed images, \textit{Sunlamp} is the most challenging test set. \textit{Lightbox-500} contains the images used for training the NeRF. Our training set contains images generated through the trained NeRF and uses the same ground-truth labels as the synthetic training set of SPEED+~\cite{park2022speed+}. Our purpose is to demonstrate that the NeRF-based training set, generated using few spaceborne images, results in a SPE network that achieves an accuracy close to the one it achieves with the large synthetic set, even if the latter approach requires the knowledge of the CAD model to generate the synthetic set while our method is model-agnostic.}
    \label{tab_datasets}
    \vspace{-0.2cm}
\end{table*}

\subsection{Offline Image Rendering}
\label{sec_appearance_embeddings}

    Training a SPE network requires a large number of images in order to capture the diversity of the pose distribution as well as of the illumination conditions.
    To generate this large training set, the learned NeRF, $m_{\Phi}$, is used to render $N_{train}$ images with pose labels randomly sampled in $\mathrm{SE}(3)$, \ie the set of rigid-body transformation in 3D space. As introduced in ~\cite{martin2021nerf}, for each image, an appearance embedding is generated by interpolating two random appearance embeddings from the NeRF training set, \ie, let $\alpha$ be a random scalar between 0 and 1, and let $\boldsymbol{e}_i$ and $\boldsymbol{e}_j$ be two randomly picked appearance embeddings from the NeRF training images, the interpolated appearance embedding $e$ is computed as:
    \begin{equation}
        \boldsymbol{e} = \boldsymbol{e}_i + \alpha (\boldsymbol{e}_j-\boldsymbol{e}_i)
    \end{equation}
    \Cref{fig_nerf_appearance_diversity} depicts several images generated using this appearance interpolation strategy.

\subsection{SPE Network Training}

    The off-the-shelf Spacecraft Pose Estimation (SPE) network, $f_{\Theta}$, is trained on the generated $N_{train}$ images. Its trained weights, $\Theta$, are finally uploaded on the chaser spacecraft.

\section{Experiments}

    This section demonstrates how our method successfully enables the training of a Spacecraft Pose Estimation (SPE) network from a sparse set of real images depicting an unknown target spacecraft. Implementation details are provided in \cref{sec_imp_details}. \cref{sec_expe_main} discusses the accuracy of the proposed approach while \cref{sec_expe_ablation} provides an ablation study evaluating the impact of the in-the-wild abilities of the NeRF on the SPE accuracy.

    \subsection{Implementation Details}
    \label{sec_imp_details}

    \subsubsection*{Dataset}
        Our method is validated on the SPEED+ dataset~\cite{park2022speed+}, used in the Spacecraft Pose Estimation Challenge (SPEC) 2021~\cite{park2023satellite}, co-organized by the European Space Agency and Stanford University. SPEED+ contains synthetic and Hardware-In-the-Loop (HIL) images of the TANGO spacecraft from the PRISMA mission~\cite{gill2007autonomous} and the corresponding pose labels. The spacecraft CAD model was not released. SPEED+ contains 59,960 grayscale synthetic images of resolution 1920x1200. The inter-spacecraft distance is between 2.2 and 10 meters. In the following experiments, these synthetic images are divided into a training set ($80\%$) and a validation set ($20\%$ ).

        The Hardware-In-the-Loop (HIL) images were taken in the Testbed for Rendezvous and Optical Navigation (TRON)~\cite{park2021robotic} facility, which emulates the lighting conditions encountered in Low Earth Orbit. HIL images are split in two domains, \textit{lightbox} and \textit{sunlamp} that aim at replicating different illumination conditions. The \textit{lightbox} domain contains 6740 images where the Earth albedo is simulated using light boxes, while the \textit{sunlamp} domain contains 2791 images where the direct illumination of the Sun is simulated using a metal halide lamp. In both domains, the inter-spacecraft distance is between 2.5 and 9.5 meters.

\begin{figure}[t]
    \centering
    \includegraphics[width=0.245\linewidth]{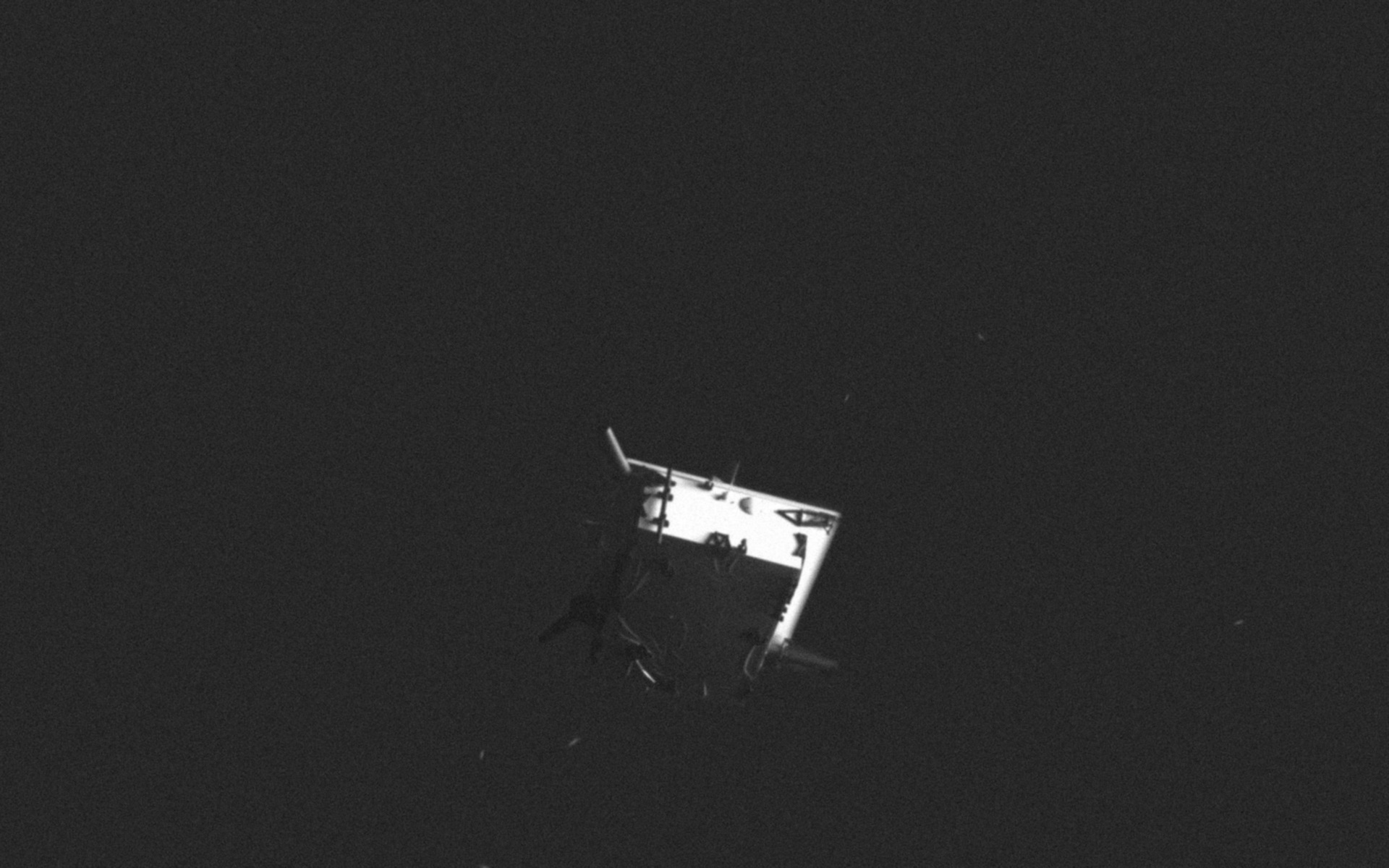}\hfill
    \includegraphics[width=0.245\linewidth]{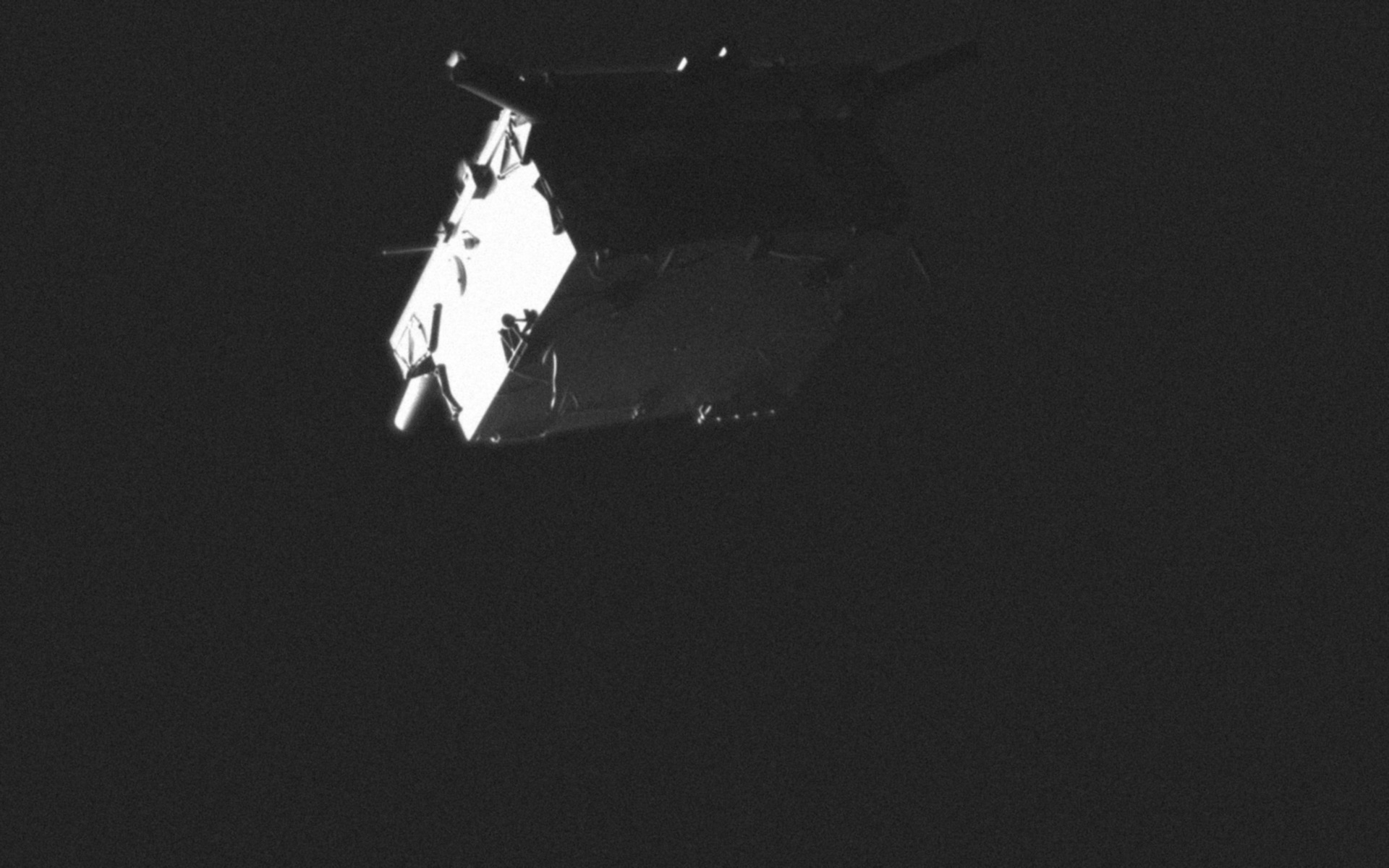}\hfill
    \includegraphics[width=0.245\linewidth]{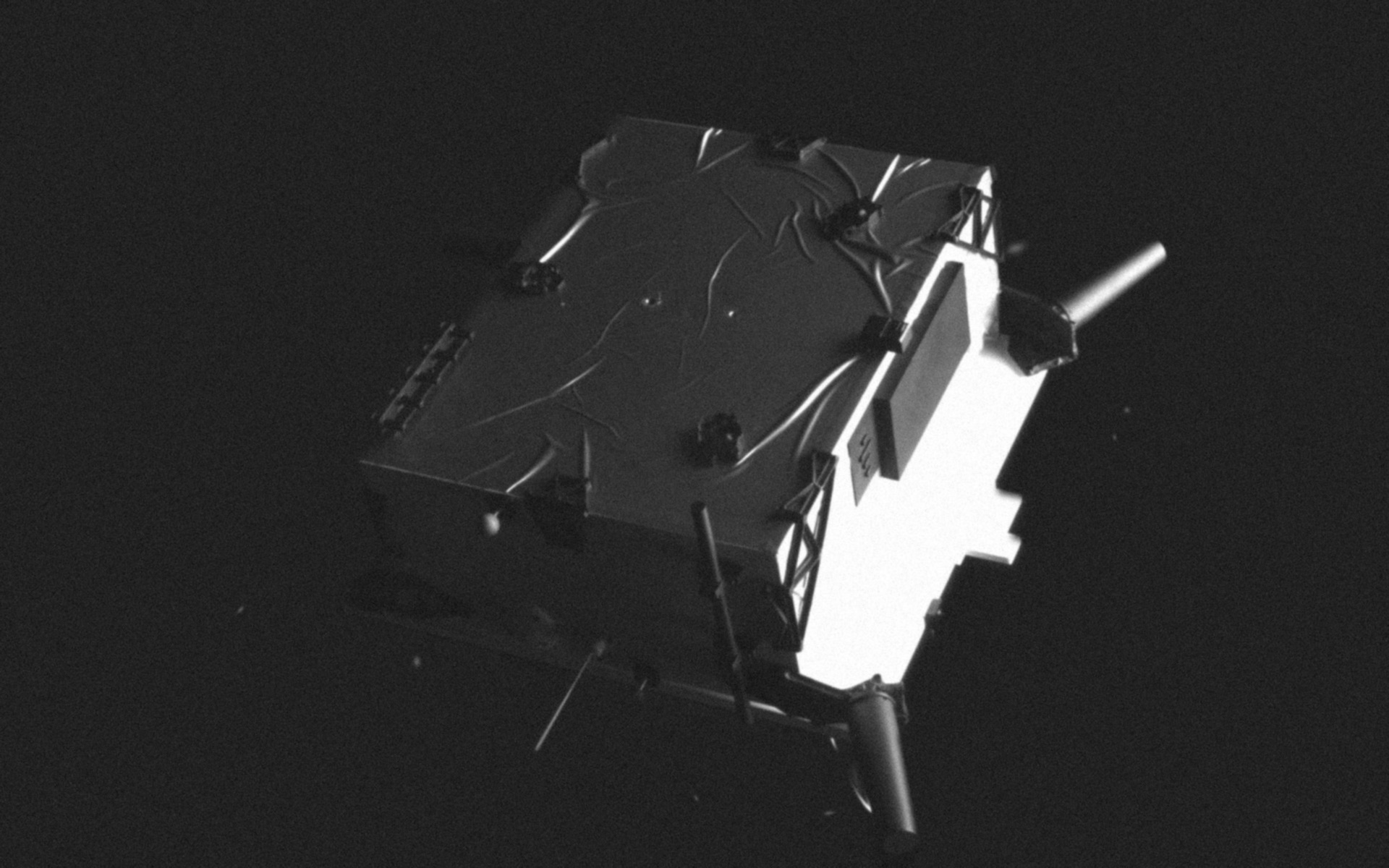}\hfill
    \includegraphics[width=0.245\linewidth]{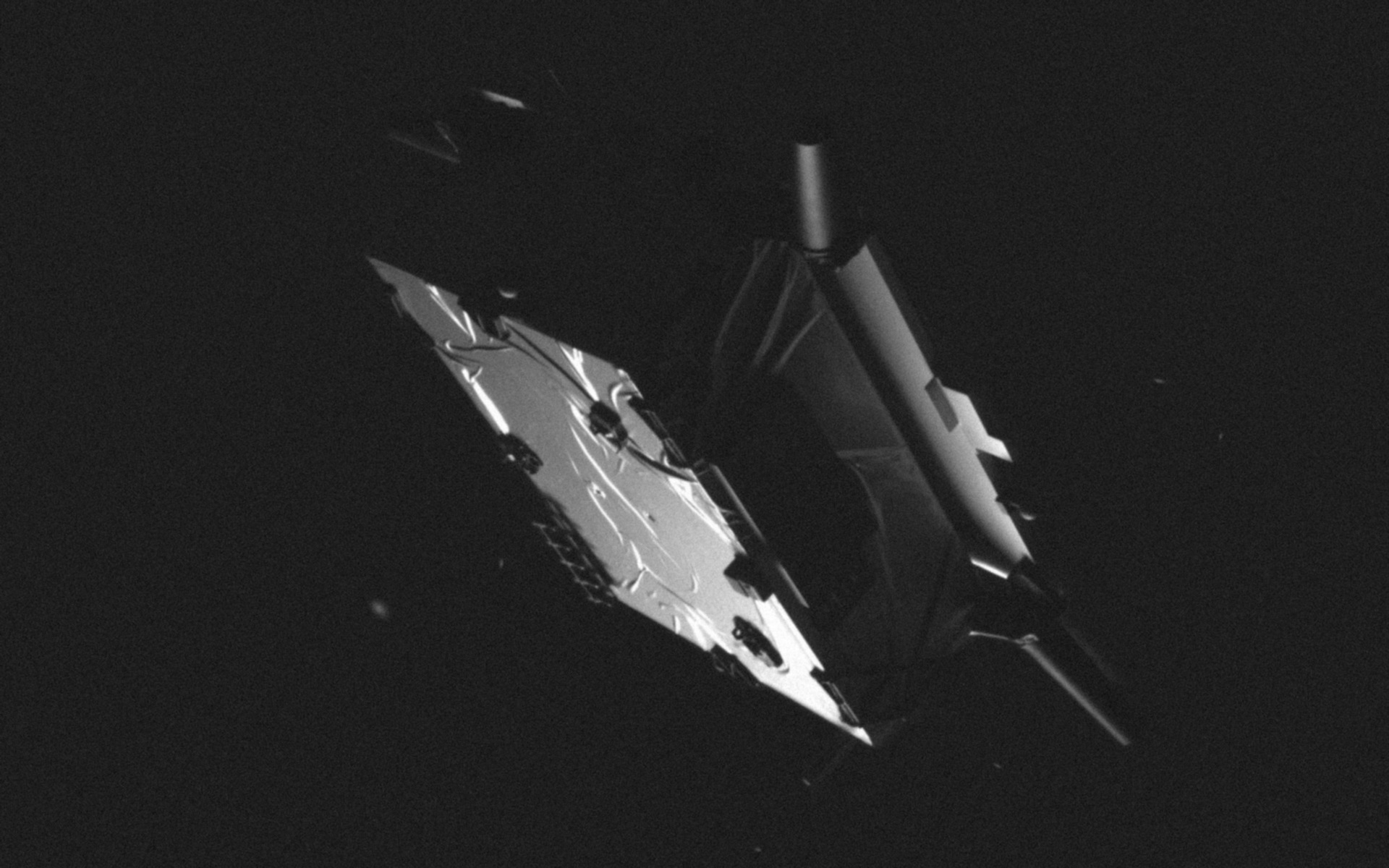}\\
    \includegraphics[width=0.245\linewidth]{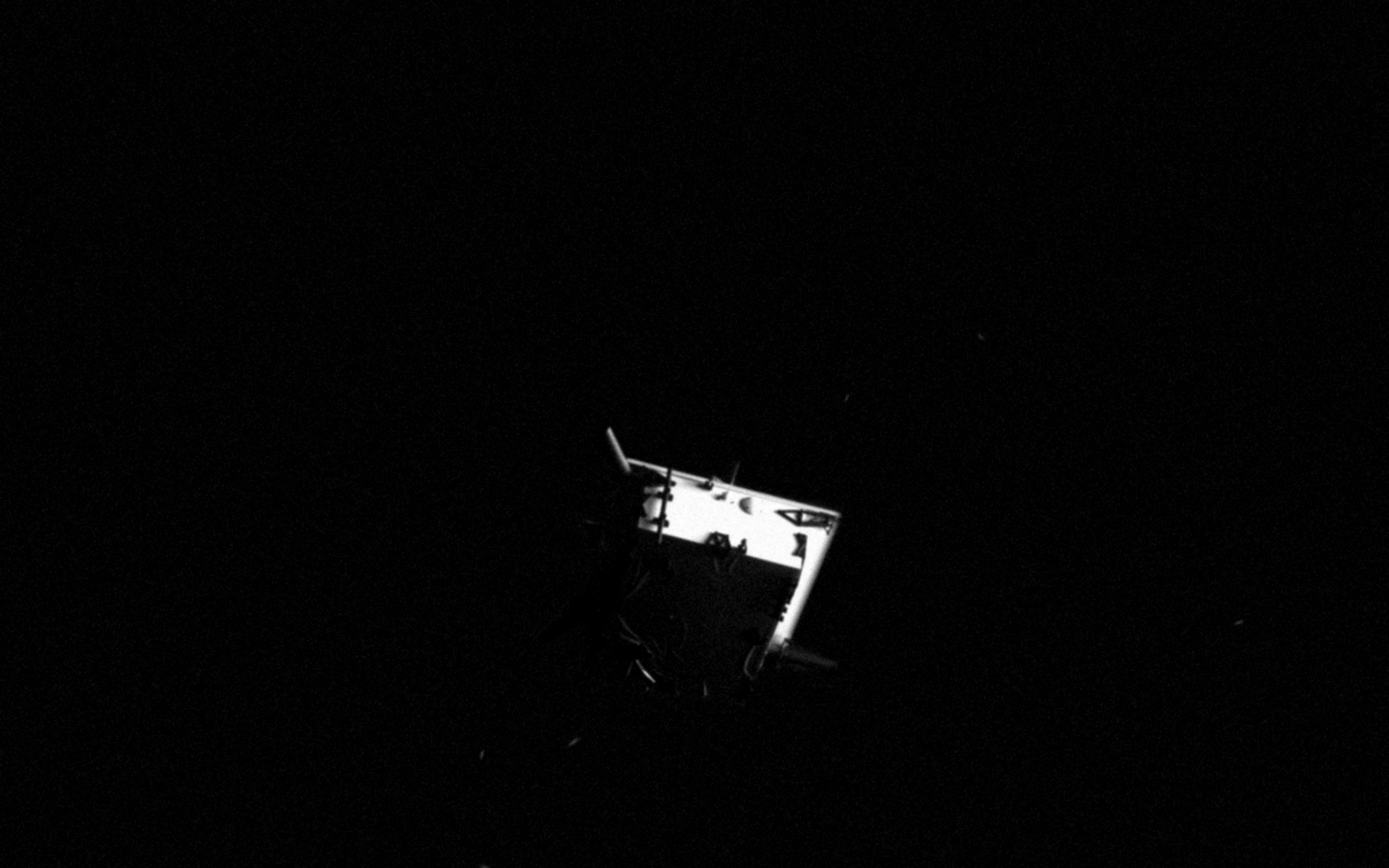}\hfill
    \includegraphics[width=0.245\linewidth]{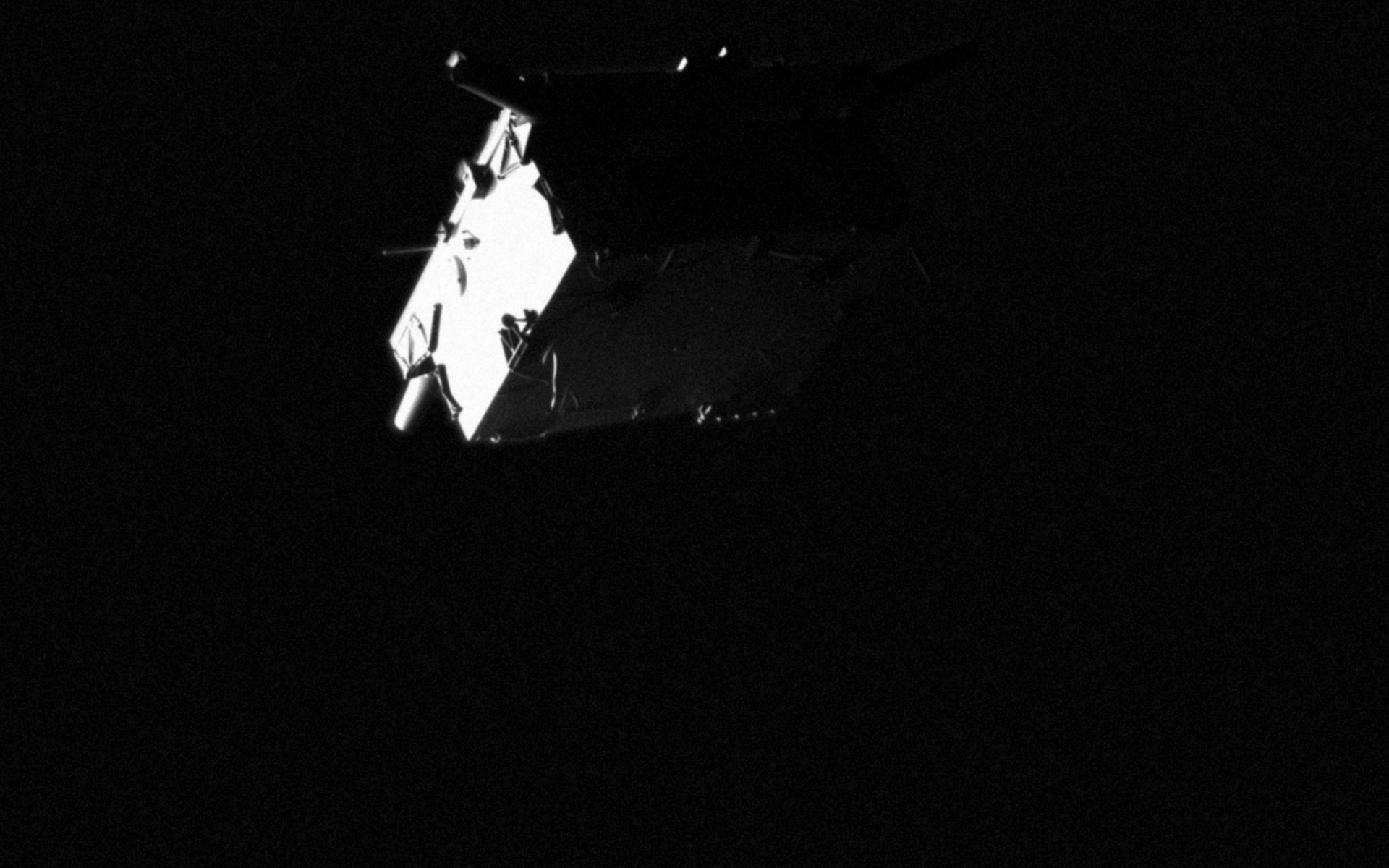}\hfill
    \includegraphics[width=0.245\linewidth]{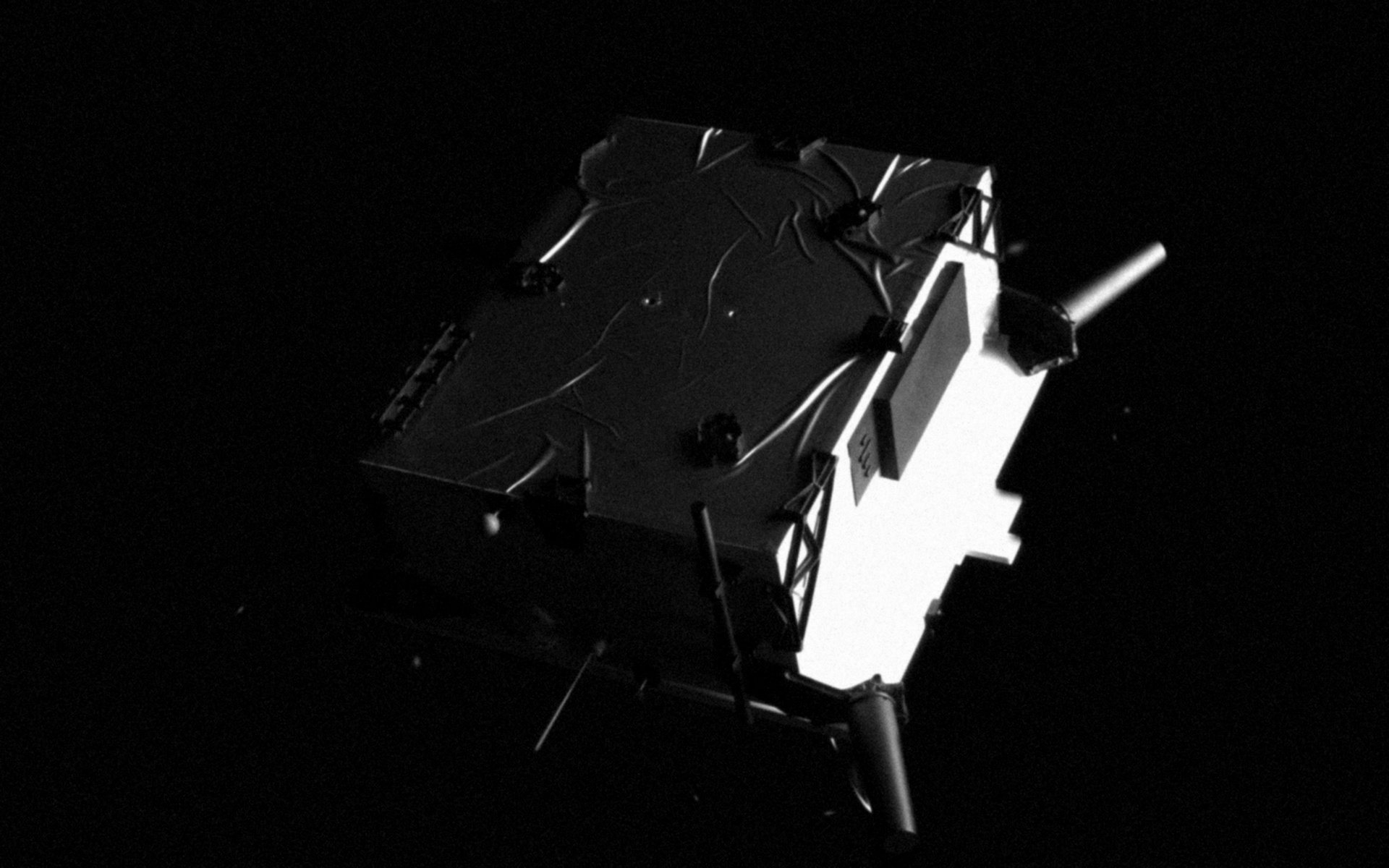}\hfill
    \includegraphics[width=0.245\linewidth]{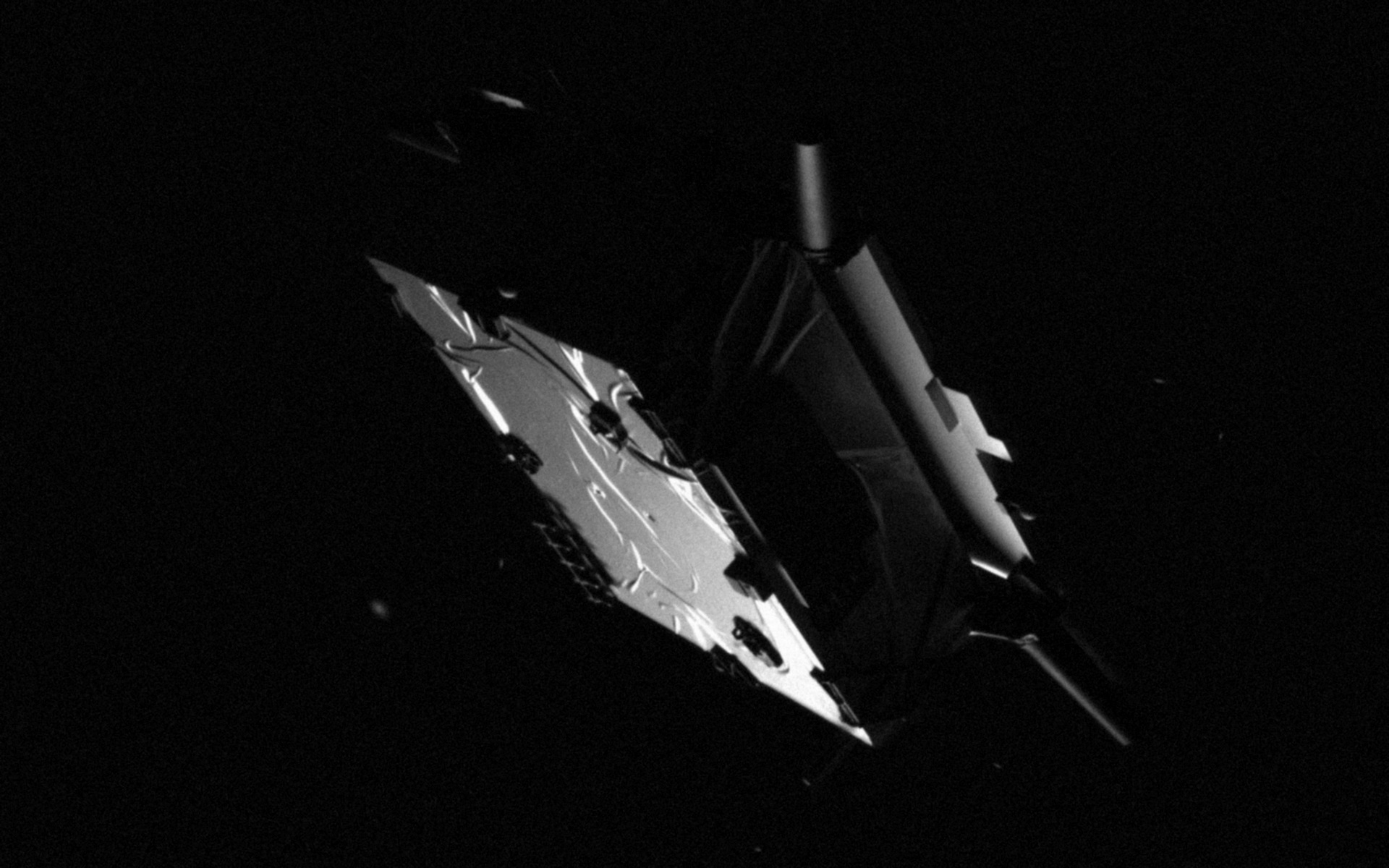} \\
    \caption{\textbf{(Top)} Original images. \textbf{(Bottom)} Images after pre-processing. For each image, the average intensity of the background is subtracted and the range is extended so that the maximal intensity of both images is unchanged.}       
    \label{fig_black_background}
    \vspace{-0.2cm}
\end{figure}
    
\begin{figure}[t]
    \centering
    \includegraphics[width=1.0\linewidth]{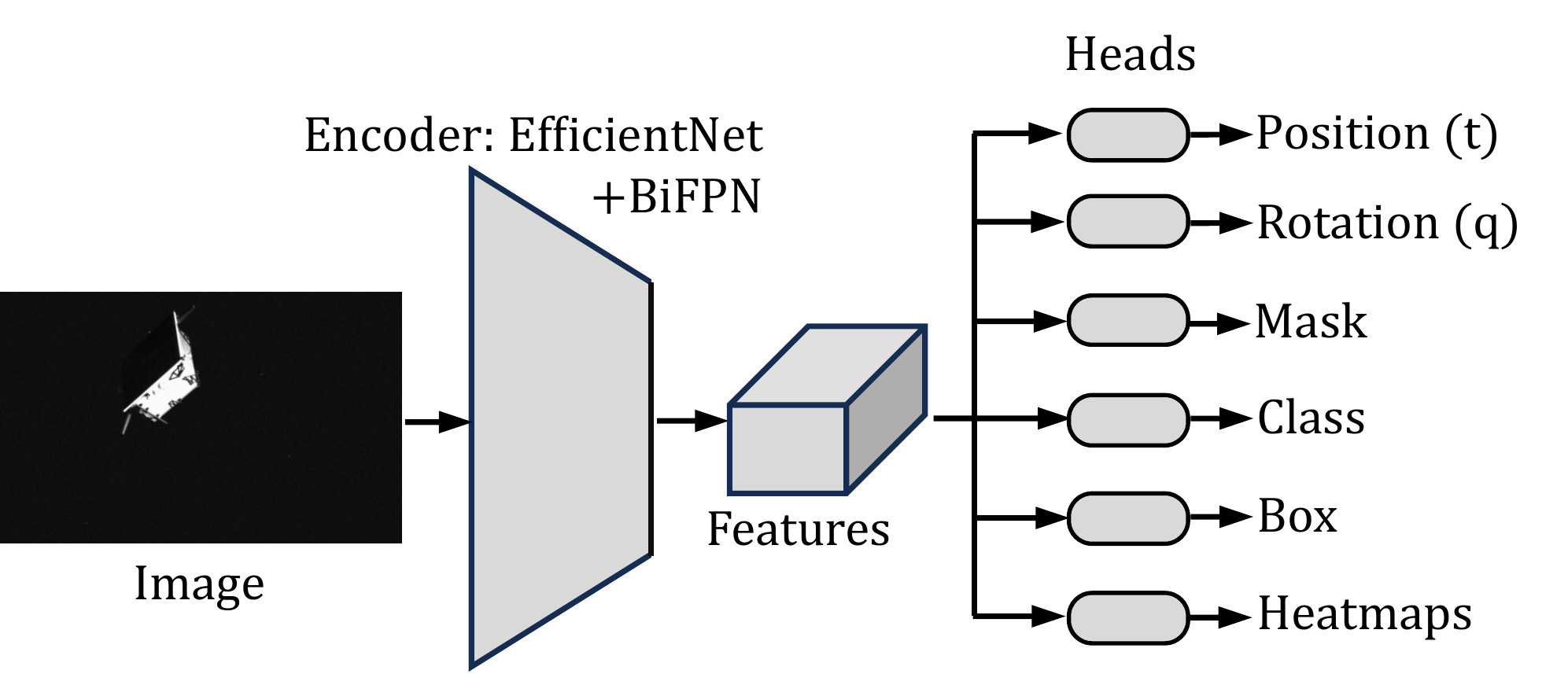}
    \vspace{-0.5cm}
    \caption{Overview of the off-the-shelf Spacecraft Pose Estimation network used in this paper, \ie SPNv2~\cite{park2023robust}. The network consists in a shared encoder followed by six heads inspired from EfficientPose~\cite{bukschat2020efficientpose}. The encoder is made of an EfficientNet backbone~\cite{tan2019efficientnet} and a BiFPN~\cite{tan2020efficientdet}. Two heads are used to estimate the relative pose, \ie position and orientation while the four last heads are used to enhance the generalization abilities of the network through Multi-Task Learning. The auxiliary tasks consist in classification, bounding box prediction, foreground segmentation and heatmap-based keypoint regression. The $K=11$ keypoints correspond to the 8 corners of the spacecraft body and the top of 3 antennas.}
    \label{fig_spnv2}
\end{figure}

        In this paper, similar to~\cite{park2023robust,park2023satellite}, the \textit{lightbox} set is assumed to accurately render the lighting conditions encountered in Low-Earth Orbit, and will thus be used to provide training and testing images that have to be representative of real images captured in orbit. From this set, all the images where (i) the inter-spacecraft distance is below five meters, or (ii) the Earth appears in the background, or (iii) the spacecraft is under-exposed or over-exposed, are removed, as explained in \cref{sec_method_selection}. 500 images are randomly picked from the remaining ones and the average intensity of their background is removed from each image, as illustrated in \Cref{fig_black_background}, to ensure a smooth training. The resulting set is named \textit{lightbox-500}. It is used only for training the NeRF while the rest of \textit{lightbox}, dubbed as \textit{lightbox*}, is kept for testing. The Neural Radiance Field, $m_{\phi}$, is trained on $N_{nerf}$ images, resized to 960x600 pixels, randomly picked from \textit{lightbox-500}. Although the pose labels associated to the $N_{nerf}$ images could be annotated through point matching across the images, we use the pose labels provided by the SPEED+ dataset~\cite{park2022speed+}. The NeRF is then used to generate a set of $N_{train}=48,000$ images to train the pose estimation model $f_{\Theta}$. Those $N_{train}$ images share the same pose labels as the synthetic train set of SPEED+. Finally, the SPE network, $f_{\Theta}$, is trained on the rendered set and the SPE metrics are computed on the \textit{sunlamp} and \textit{lightbox*} sets. \Cref{tab_datasets} presents an overview of the sets used in this paper.

    \subsubsection*{In-the-wild NeRF $m_{\Phi}$}
    
     In this paper, the in-the-wild Neural Radiance Field, $m_{\Phi}$, follows the $K$-Planes~\cite{fridovich2023k} implementation because it features both learnable appearance embeddings and efficient encoding, thereby reducing the training time. $m_{\Phi}$ uses the default configuration of $K$-Planes except that it does not use a linear decoder but a MLP~\cite{fridovich2023k}. The appearance embedding dimension is kept at 32. The network is trained for 30,000 steps of 4096 points per batch. The NeRF training takes 40 minutes on a NVIDIA RTX3090. \Cref{fig_nerf_appearance_diversity} highlights the ability of $K$-Planes to render not only the pose distribution, but also variable illumination conditions thanks to its appearance embedding. $m_{\Phi}$ is then used to render 48,000 images that correspond to the pose labels from the synthetic training set of SPEED+~\cite{park2022speed+} used with randomly interpolated appearance embeddings, as explained in \cref{sec_appearance_embeddings}. The rendering of the whole set takes 16 GPU-hours and is performed on a cluster made of 4 NVIDIA RTX3090 and 10 NVIDIA TeslaA100 GPUs.

\begin{figure}[t]
    \includegraphics[width=.25\linewidth]{NeRF_outputs/images_appearance/003_1.jpg}\hfill
    \includegraphics[width=.25\linewidth]{NeRF_outputs/images_appearance/232_1.jpg}\hfill
    \includegraphics[width=.25\linewidth]{NeRF_outputs/images_appearance/295_1.jpg}\hfill
    \includegraphics[width=.25\linewidth]{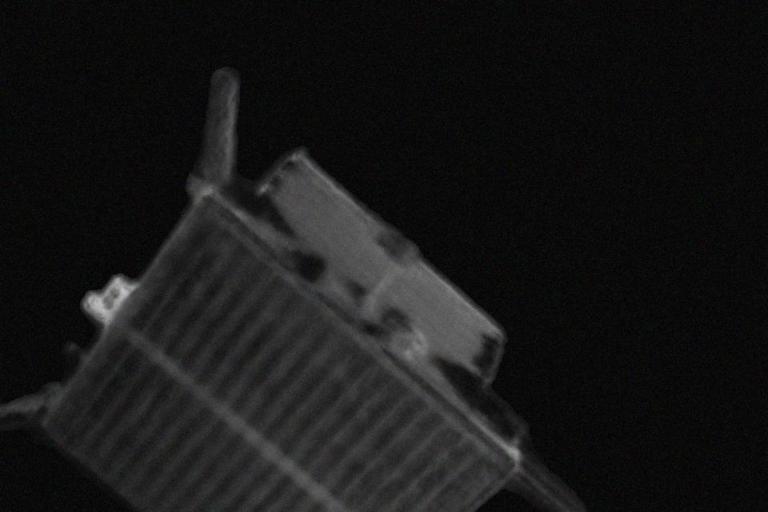}\\    
    \includegraphics[width=.25\linewidth]{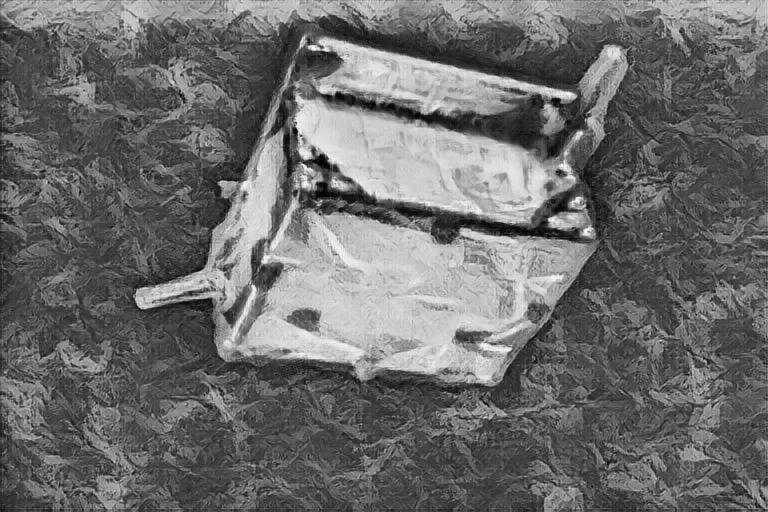}\hfill    
    \includegraphics[width=.25\linewidth]{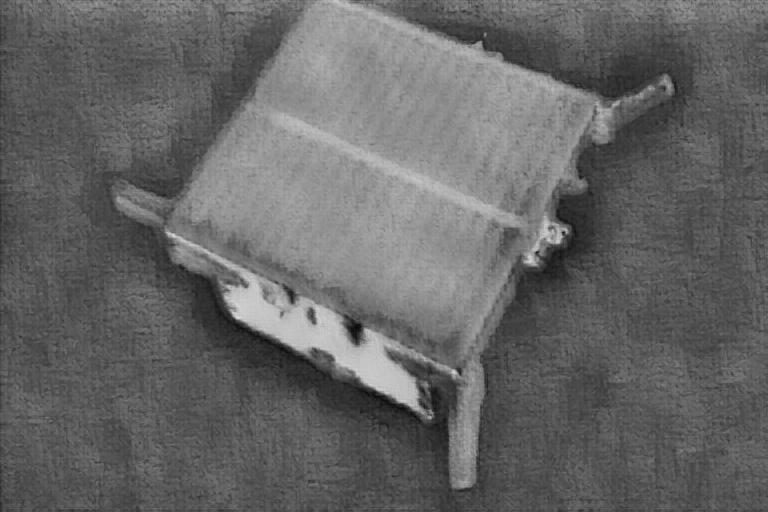}\hfill    
    \includegraphics[width=.25\linewidth]{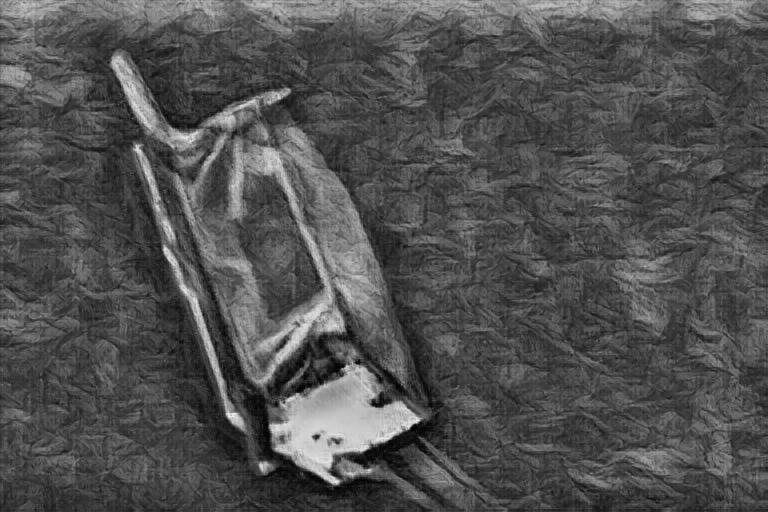}\hfill    
    \includegraphics[width=.25\linewidth]{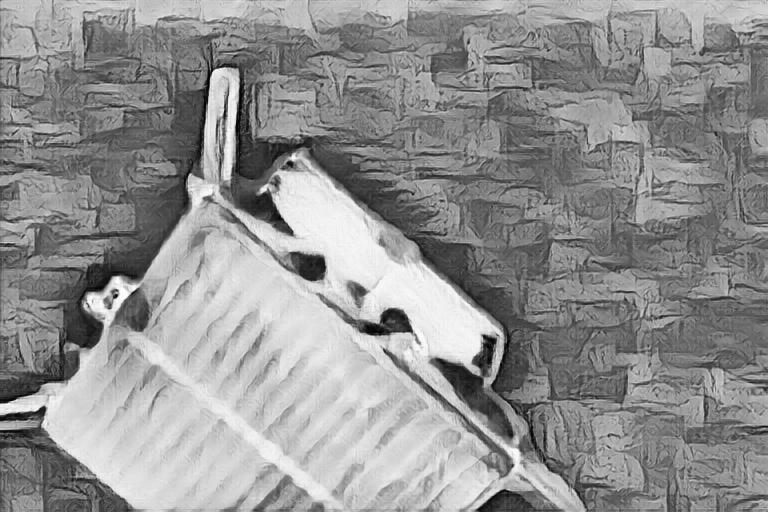}\\
    \caption{Examples of Neural Style Augmentation~\cite{jackson2019style}. \textbf{(Top)} Original images \textbf{(Bottom)} Corresponding images after Style Augmentation. The spacecraft shape is always preserved by the transformation but its texture is augmented.}
    \label{fig_style_images}
    \vspace{-0.2cm}
\end{figure}

\begin{figure}[t]
    \centering
    \includegraphics[width=0.245\linewidth]{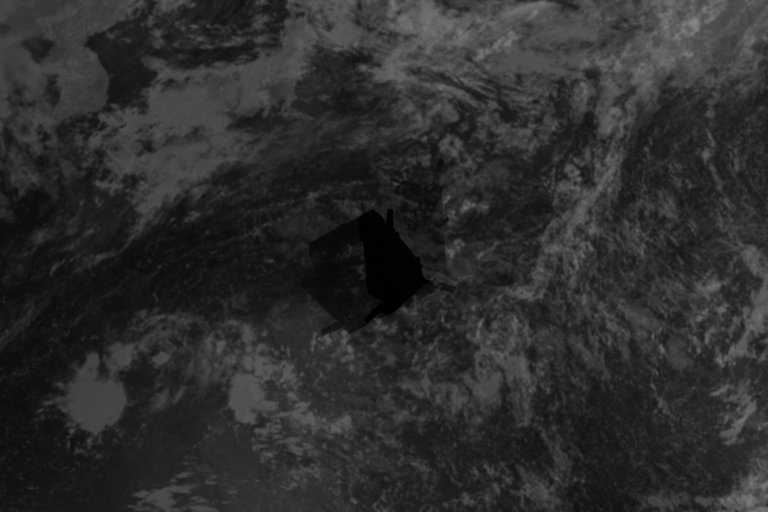}\hfill
    \includegraphics[width=0.245\linewidth]{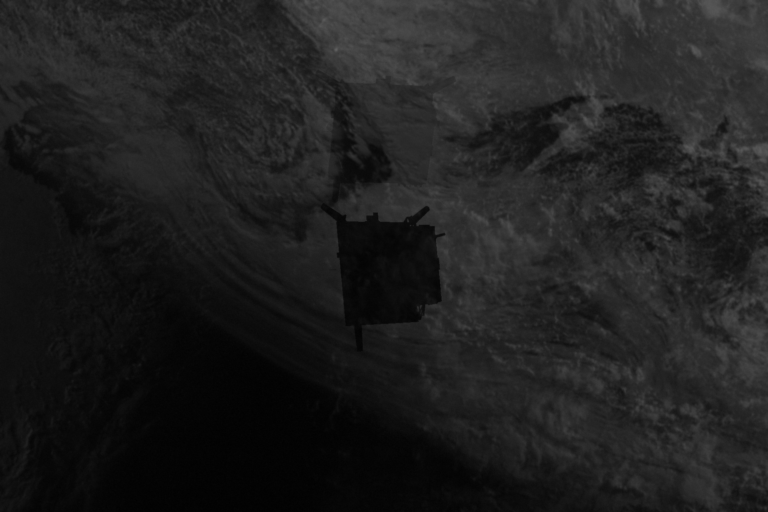}\hfill
    \includegraphics[width=0.245\linewidth]{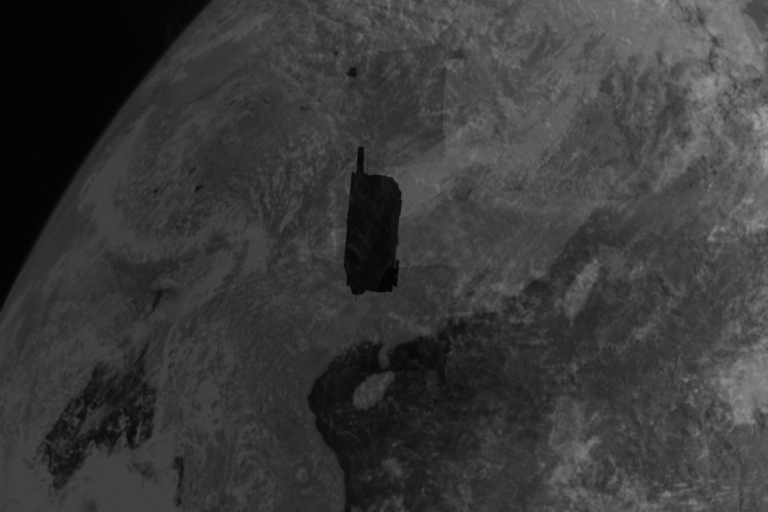}\hfill
    \includegraphics[width=0.245\linewidth]{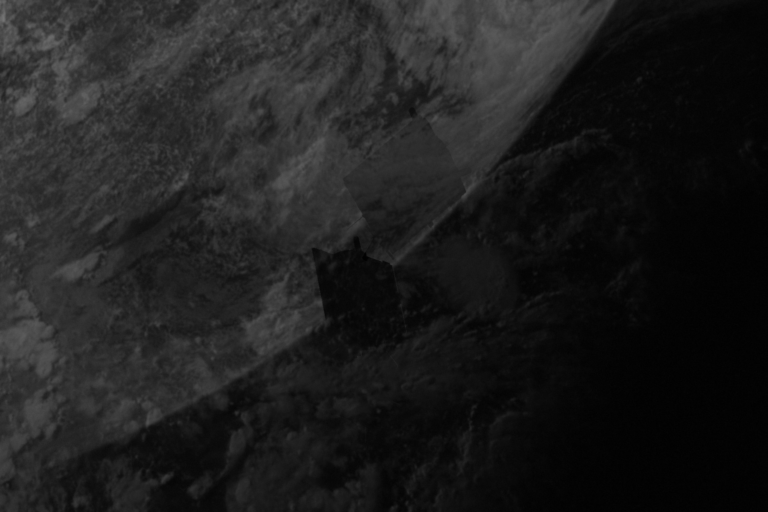}\\   
    \includegraphics[width=0.245\linewidth]{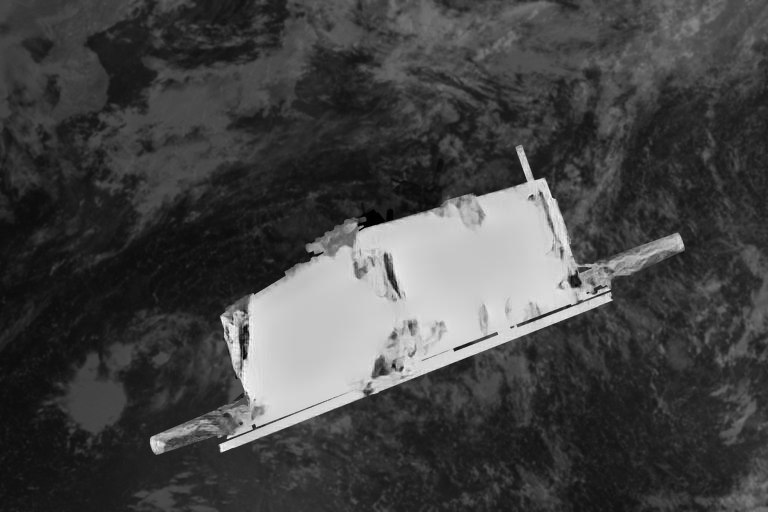}\hfill      
    \includegraphics[width=0.245\linewidth]{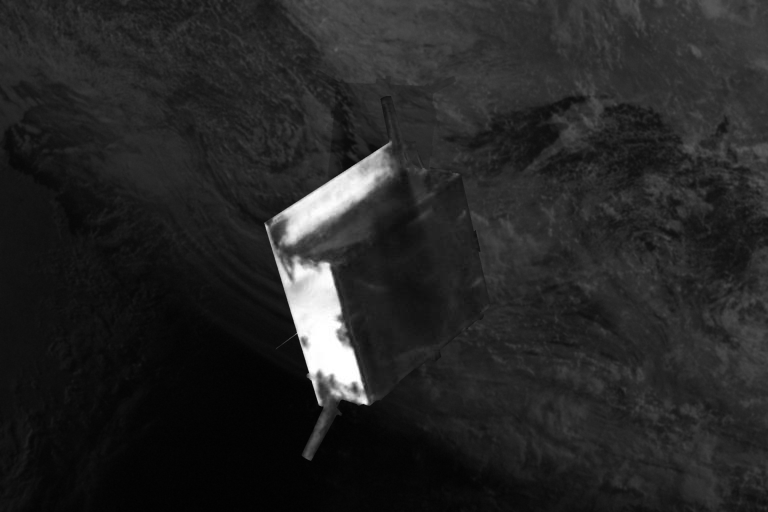}\hfill  
    \includegraphics[width=0.245\linewidth]{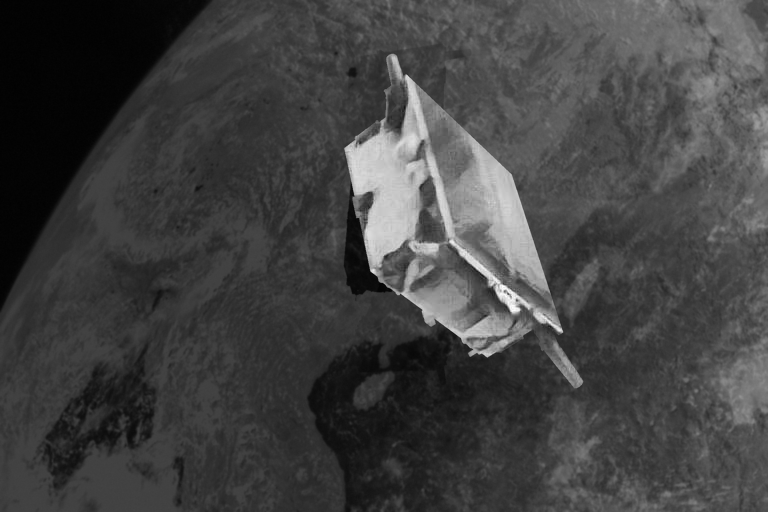}\hfill
    \includegraphics[width=0.245\linewidth]{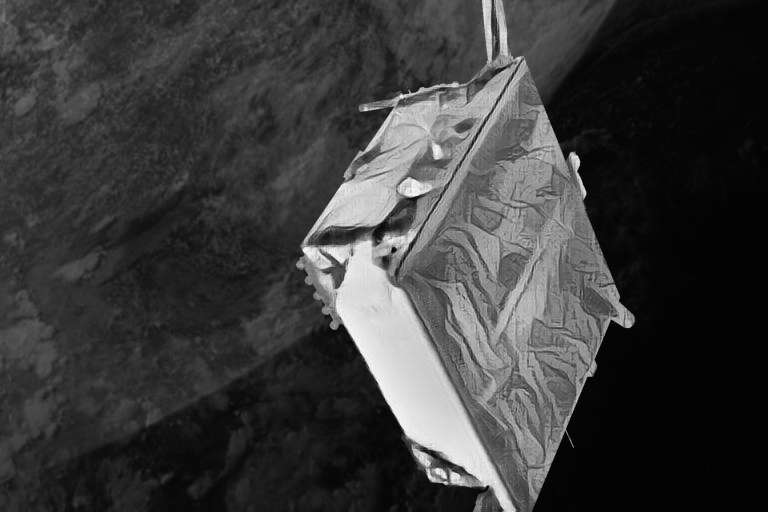}\\ 
    \caption{\textbf{(Top)} Examples of background images recovered by merging 4 randomly picked background images from the SPEED+~\cite{park2022speed+} synthetic set. \textbf{(Bottom)} Examples of images augmented with these backgrounds. The two first ones are performed from the NeRF output while the two last ones first went through the style augmentation stage.}    
    \label{fig_earth_background}
    \vspace{-0.2cm}
\end{figure}

\begin{table*}[t]
\centering
\begin{tabular}{c|r|c|rrrr|rrrr}
    \toprule
    Training  Strategy & $N$ & No CAD  & \multicolumn{4}{c}{\textit{Sunlamp}} & \multicolumn{4}{c}{\textit{Lightbox*}} \\
     & & required & $S^{*}_{Pose} [/]$ & $E_{R}[\degree$] & $E_{T}[m]$ & $\bar{E}_{T}[\%]$ & $S^{*}_{Pose}[/]$ & $E_{R}[\degree]$ & $E_{T}[m]$ & $\bar{E}_{T}[\%]$ \\ 
    \midrule
    Synthetic (w. CAD) & / & & \textbf{0.322} & 15.7 & 0.30 & 0.05 & 0.203 & 9.5 & 0.25 & 0.04 \\
    \hline
    Baseline & 500  & \ding{51} & 2.005 & 99.4 & 1.20 & 0.27 & 1.96 & 95.0 & 1.31 & 0.30 \\
    NeRF (\textbf{Ours})& 50  & \ding{51} & 0.502 & 24.1 & 0.57 & 0.08 & 0.213 & 9.9 & 0.24 & 0.04 \\
    NeRF (\textbf{Ours})& 100  & \ding{51} & 0.491 & 23.6 & 0.54 & 0.08 & 0.161 & 7.4 & 0.19 & 0.03 \\
    NeRF (\textbf{Ours})& 200  & \ding{51} & 0.375 & \textbf{15.3} & 0.37 & 0.06 & \textbf{0.158} & \textbf{7.2} & \textbf{0.20} & \textbf{0.03}\\
    NeRF (\textbf{Ours})& 500  & \ding{51} & 0.341 & 16.9 & \textbf{0.29} & \textbf{0.04} & \textbf{0.158} & \textbf{7.2} & \textbf{0.20} & \textbf{0.03}\\
    \bottomrule
\end{tabular}
\caption{Comparison of three SPE training strategies, \ie, (i) trained on 48,000 synthetic images generated using the CAD model of the target, (ii) trained on $N$ real images (baseline), or (iii) trained on 48,000 images generated by a NeRF trained on $N$ real images, and tested on two test sets, \ie, \textit{Sunlamp} and \textit{Lightbox*}. Our method significantly outperforms the baseline and reaches similar performance as the synthetic approach while requiring no CAD model of the target.}
\label{tab_main_results}
\vspace{-0.2cm}
\end{table*}

\begin{figure}[t]
    \includegraphics[width=.94\linewidth]{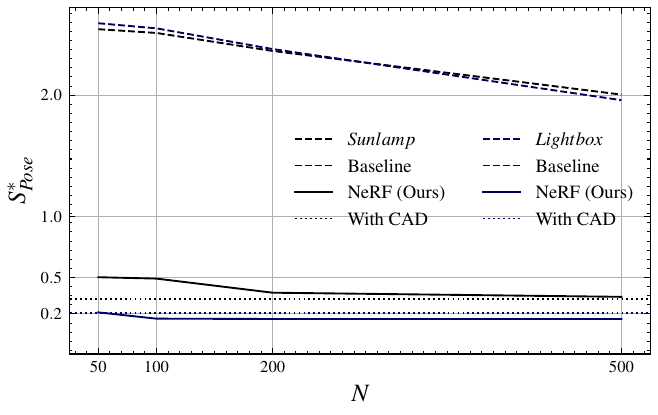}
    \vspace{-0.3cm}
    \caption{Relationship between the number of real images, $N$ and the SPEED+ score of the SPE when trained on (i) $N$ real images (dashed lines), (ii) 48,000 images generated by a NeRF trained on $N$ images (solid lines) or (iii) 48,000 synthetic images rendered with the CAD model of the target (dotted lines), for the \textit{Sunlamp} (blue) and \textit{Lightbox*} (black) sets. Our NeRF-based approach always outperforms the baseline by a large margin. When trained with enough images, \eg 500, it performs similarly as the CAD-based approach while requiring no CAD model.}
    \label{fig_graph_number_training_images}
    \vspace{-0.2cm}
\end{figure}

    \subsubsection*{Off-the-shelf SPE network $f_{\Theta}$}

    In our experiments, the Spacecraft Pose Estimation network, $f_{\Theta}$, consists in a SPNv2~\cite{park2023robust} with scaling coefficient $\phi = 3$ and Batch-Normalization layers~\cite{ioffe2015batch}. \Cref{fig_spnv2} illustrates the SPNv2 architecture. The training is conducted using the same procedure as in the original paper~\cite{park2023robust}, including the offline neural Style Augmentation ~\cite{jackson2019style}, as illustrated in \Cref{fig_style_images}. In addition, we add a custom background augmentation, illustrated in \Cref{fig_earth_background}, which is applied on half of the training images to add the Earth in background. The SPE is trained for 30,000 steps on batches of 32 images on a NVIDIA TeslaA100 GPU.

\begin{figure}[t]
    \includegraphics[width=.94\linewidth]{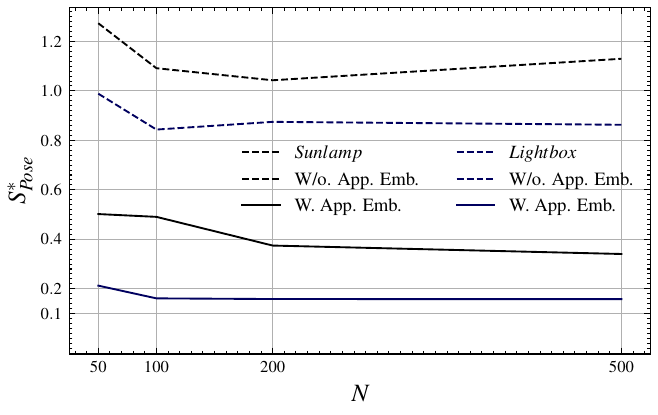}
    \caption{Relationship between the number of real images, $N$, and the SPEED+ score of the SPE network when trained on 48,000 images generated using either (i) the average appearance embedding (dashed lines) , or (ii) interpolated appearance embeddings (solid lines), trained on $N$ real images and tested on the \textit{Sunlamp} (blue) and \textit{Lightbox*} (black) sets. The SPE network trained on images rendered with interpolated appearance embeddings always outperforms by a large margin a SPE network trained on images generated with the average appearance embedding.}
    \label{fig_graph_in_the_wild}
    \vspace{-0.2cm}
\end{figure}

    \subsubsection*{Metrics}
    
    Our method is evaluated by the SPEED+ score, $S^{*}_{Pose}$, introduced in SPEC2021~\cite{park2023satellite}, which averages the pose scores of the N images of the test set. For each image, the translation error, $e_{t}$, is computed as the norm of the error between the predicted position, $\hat{t}$, and the ground-truth position, $t$, \ie
    \begin{equation}
        e_{t} = \left |\left | \hat{t} - t  \right |\right |,
    \end{equation}
    while the normalized translation error, $\bar{e}_{t}$, is equal to the ratio between the translation error and the norm of the ground-truth position, \ie
    \begin{equation}
        \bar{e}_{t} = \frac{e_{t}}{\left |\left |  t  \right |\right |}.
    \end{equation}
    The rotation error, $e_{q}$, is computed as the angular error between the predicted quaternion, $\hat{q}$, and the ground-truth quaternion, $q$, \ie
    \begin{equation}
        e_{q} = 2 \textnormal{arccos} \left( \left| \hat{q} q^{T}\right| \right).
    \end{equation}

    As the ground-truth positions and rotations of the HIL domains were measured with some uncertainty~\cite{park2022speed+}, the predictions are considered as perfect if they are lower than the calibration error, \ie
    \begin{equation}
      e^{*}_{q} = 
      \begin{cases}
        0, & \text{if } e_{q} < 0.00295 \textnormal{rad}\\
        e_{q}, & \text{otherwise},
      \end{cases}
    \end{equation}
    and
    \begin{equation}
      \bar{e}^{*}_{t} = 
      \begin{cases}
        0, & \text{if } \bar{e}_{t} < 2.173 \textnormal{mm/m}\\
        \bar{e}_{t}, & \text{otherwise}
      \end{cases}
    \end{equation}

\begin{table*}[t]
\centering
\begin{tabular}{c|rrrr|rrrr}
    \toprule
    Offline Rendering  & \multicolumn{4}{c}{\textit{Sunlamp}} & \multicolumn{4}{c}{\textit{Lightbox*}} \\
      & $S^{*}_{Pose} [/]$ & $E_{R}[\degree$] & $E_{T}[m]$ & $\bar{E}_{T}[/]$ & $S^{*}_{Pose}[/]$ & $E_{R}[\degree]$ & $E_{T}[m]$ & $\bar{E}_{T}[/]$ \\    
    \midrule
    Average Appearance Embedding & 1.130 & 60.2 & 0.45 & 0.08 & 0.863 & 39.6 & 0.95 & 0.18 \\
    \textbf{Interpolated Appearance Embedding} & \textbf{0.341} & \textbf{16.9} & \textbf{0.29} & \textbf{0.04} & \textbf{0.158} & \textbf{7.2} & \textbf{0.20} & \textbf{0.03}\\
    \bottomrule
\end{tabular}
\caption{Comparison of the SPE metrics when trained on 48,000 images generated by a NeRF using either (i) the average appearance embedding, or (ii) using interpolated appearance embeddings. In both cases, the NeRF is trained on 500 real images. Generating the SPE training set with interpolated appearance embeddings decreases the SPE errors.}
\label{tab_results_appearance}
\end{table*}

    The average normalized translation, $E_{T,N}$, translation, $E_{T}$, and rotation errors, $E_{R}$, are computed as the average of the corresponding values over the set, \ie
    \begin{equation}
        \bar{E}_{T} =  \frac{1}{N} \sum_{i=1}^{N} \bar{e}^{(i)}_{t},
    \end{equation}   
    \begin{equation}
        E_{T} =  \frac{1}{N} \sum_{i=1}^{N} e^{(i)}_{t}, \qquad E_{R} =  \frac{1}{N} \sum_{i=1}^{N} e^{(i)}_{q}.
    \end{equation}        
    Finally, the SPEED+-score, $S^{*}_{Pose}$, computes the average of the sum of the normalized translation and rotation errors over the set, \ie
    \begin{equation}
        S^{*}_{Pose} =  \frac{1}{N} \sum_{i=1}^{N} \left( \bar{e}^{*(i)}_{t} + e^{*(i)}_{q} \right).
    \end{equation}

\begin{figure}[t]
    \includegraphics[width=.245\linewidth]{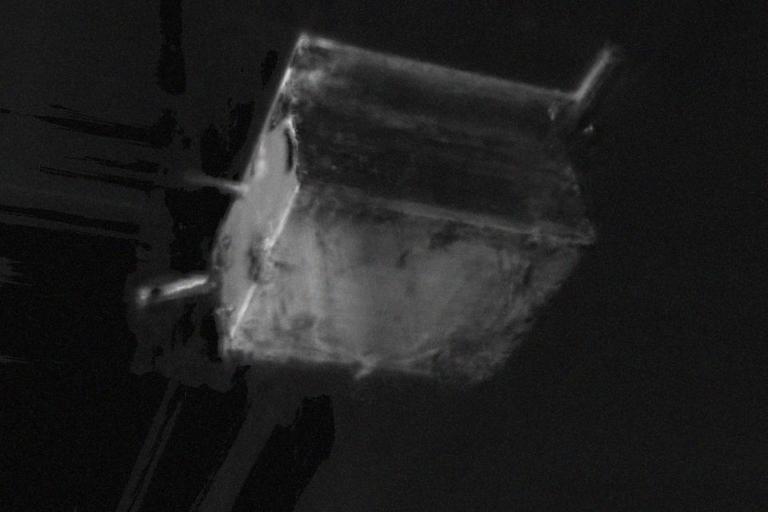}\hfill
    \includegraphics[width=.245\linewidth]{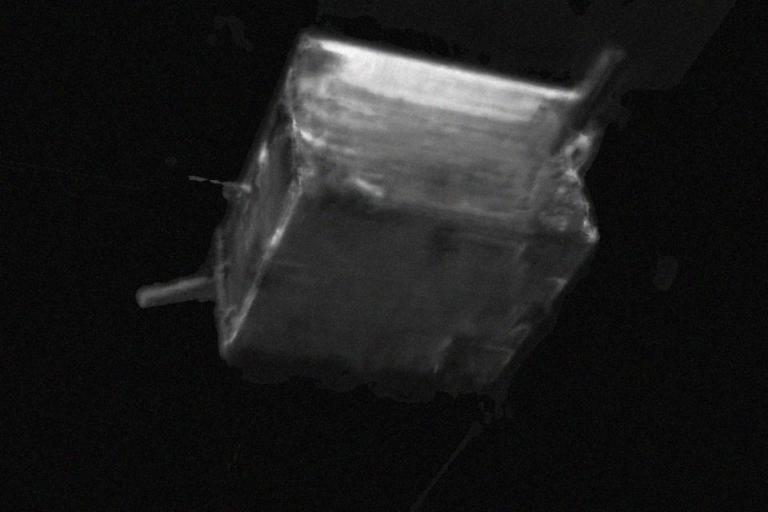}\hfill
    \includegraphics[width=.245\linewidth]{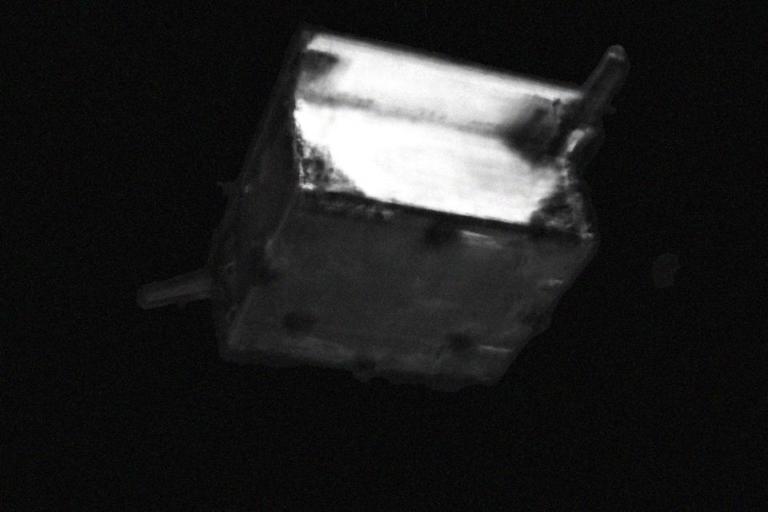}\hfill
    \includegraphics[width=.245\linewidth]{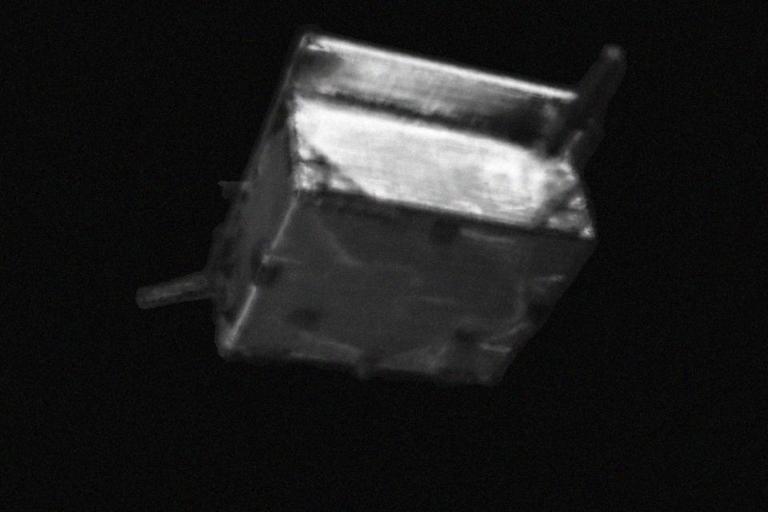}\\    
    \includegraphics[width=.245\linewidth]{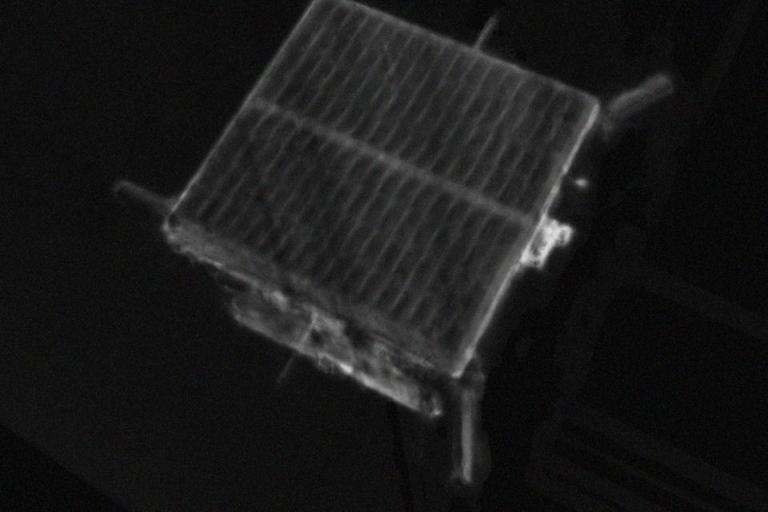}\hfill
    \includegraphics[width=.245\linewidth]{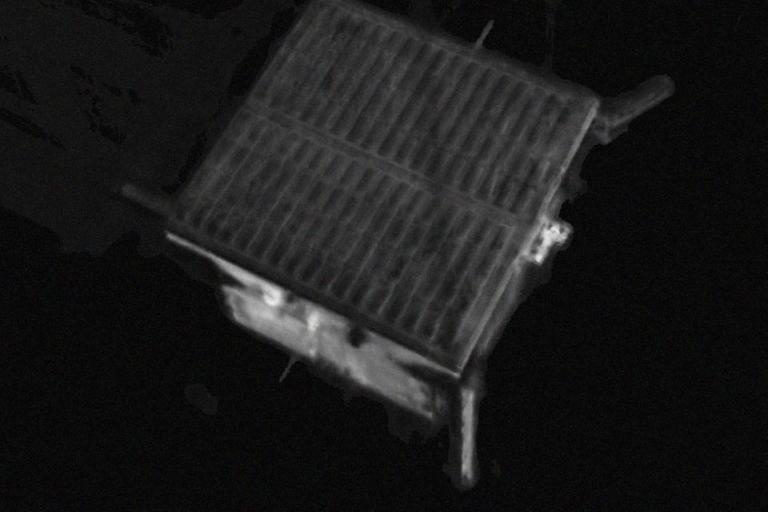}\hfill
    \includegraphics[width=.245\linewidth]{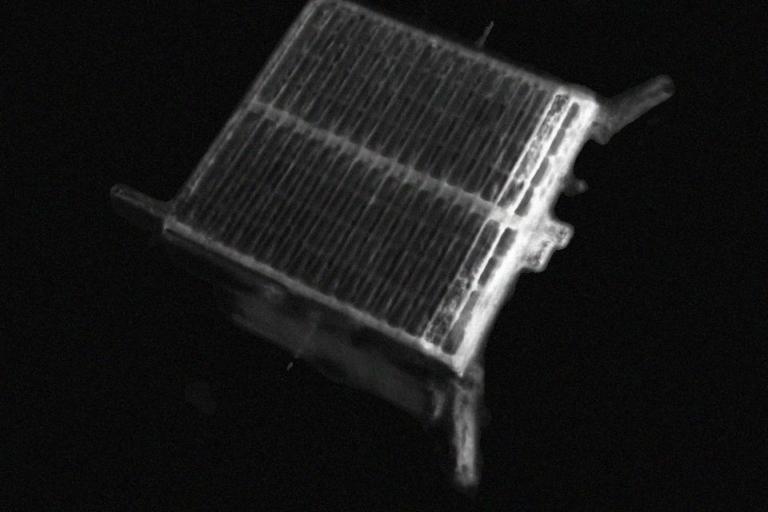}\hfill
    \includegraphics[width=.245\linewidth]{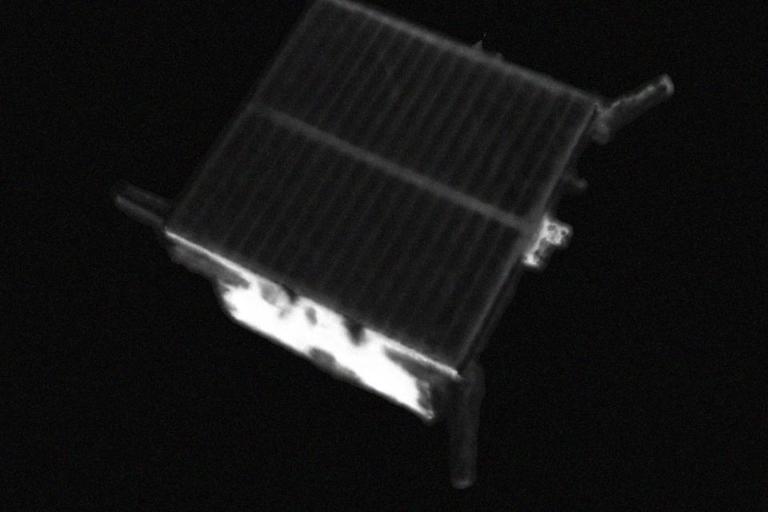}\\    
    \includegraphics[width=.245\linewidth]{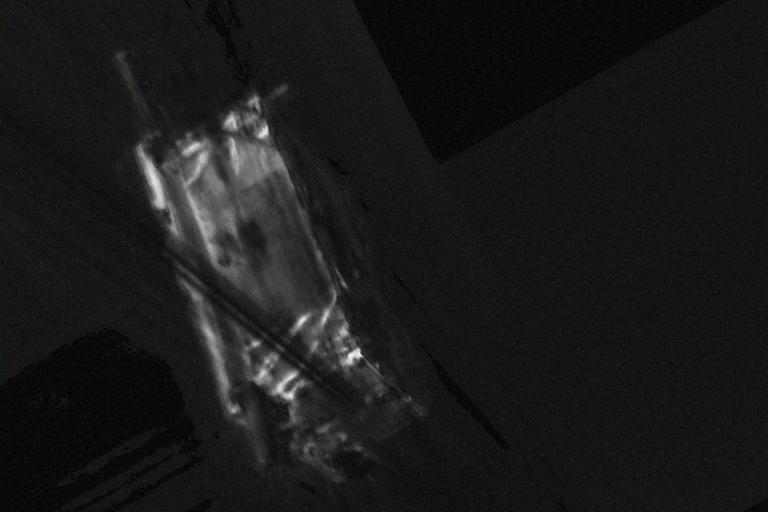}\hfill
    \includegraphics[width=.245\linewidth]{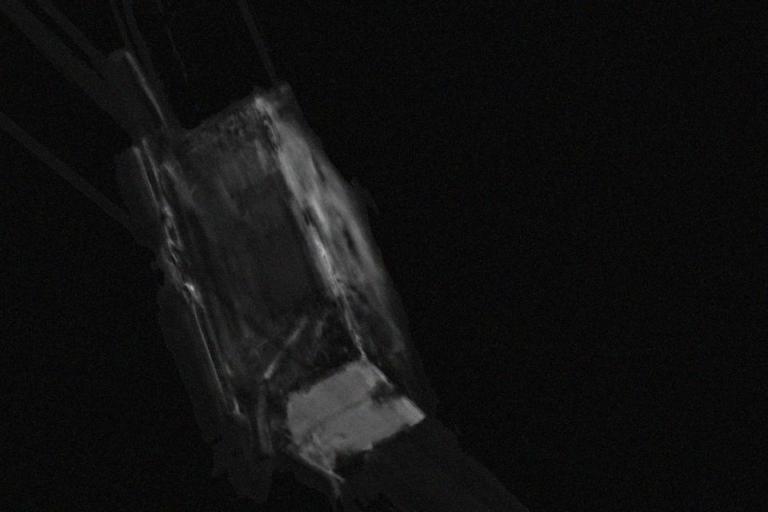}\hfill
    \includegraphics[width=.245\linewidth]{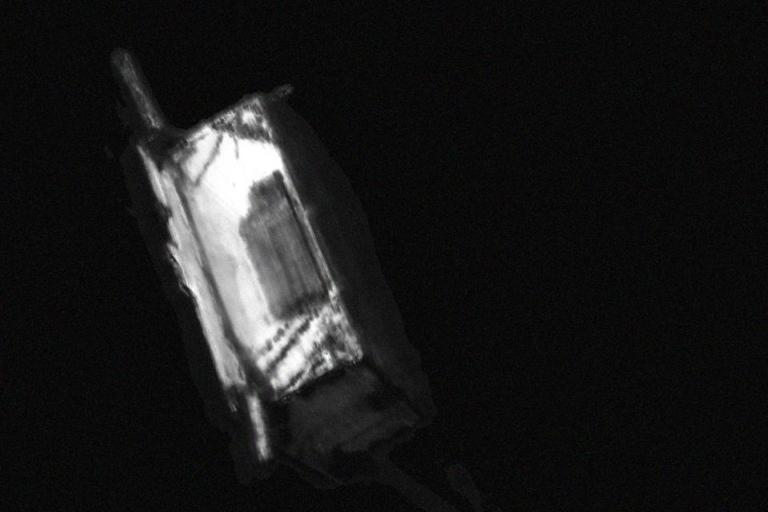}\hfill
    \includegraphics[width=.245\linewidth]{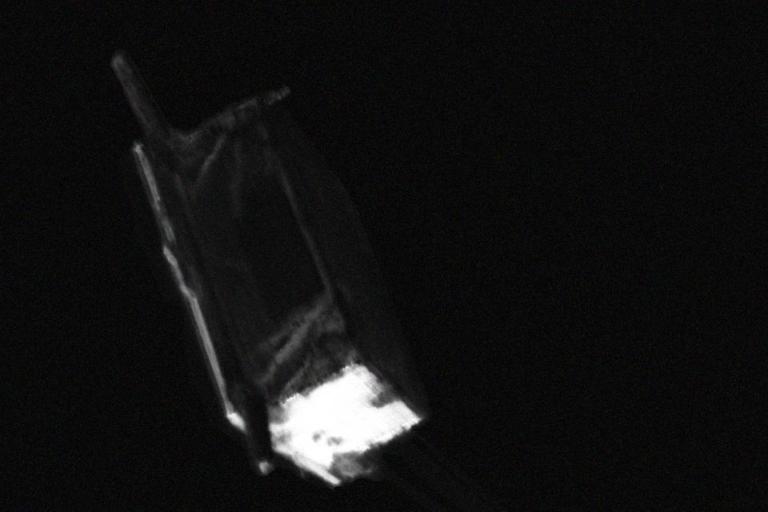}\\    
    \includegraphics[width=.245\linewidth]{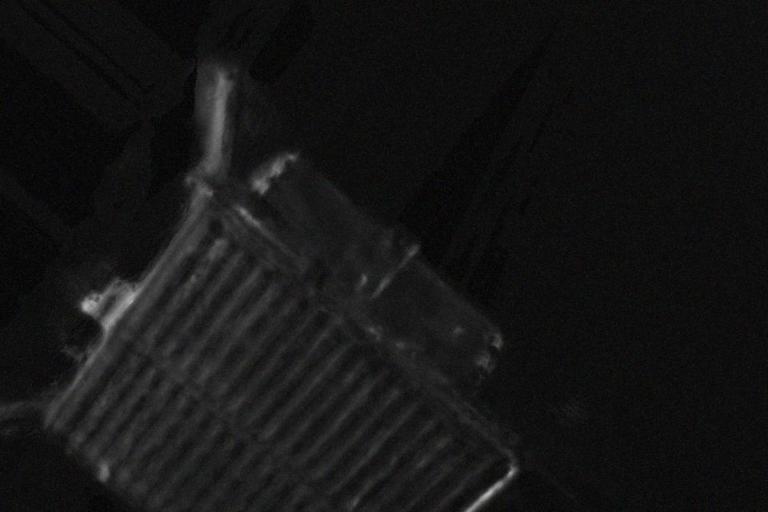}\hfill
    \includegraphics[width=.245\linewidth]{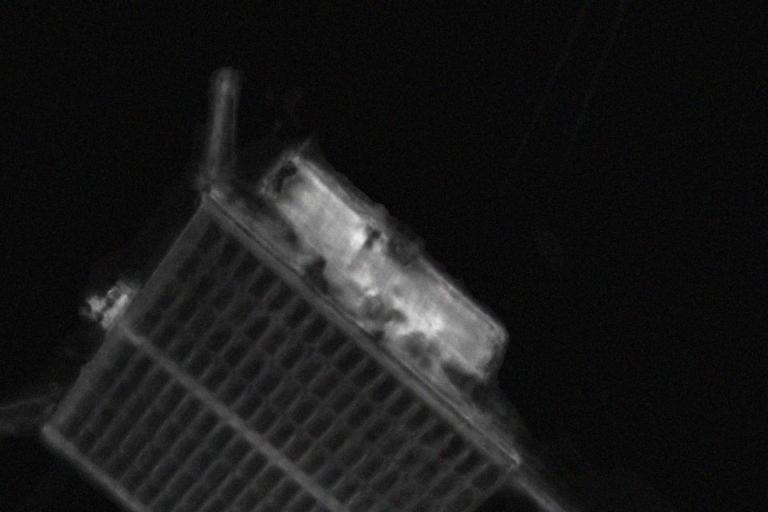}\hfill
    \includegraphics[width=.245\linewidth]{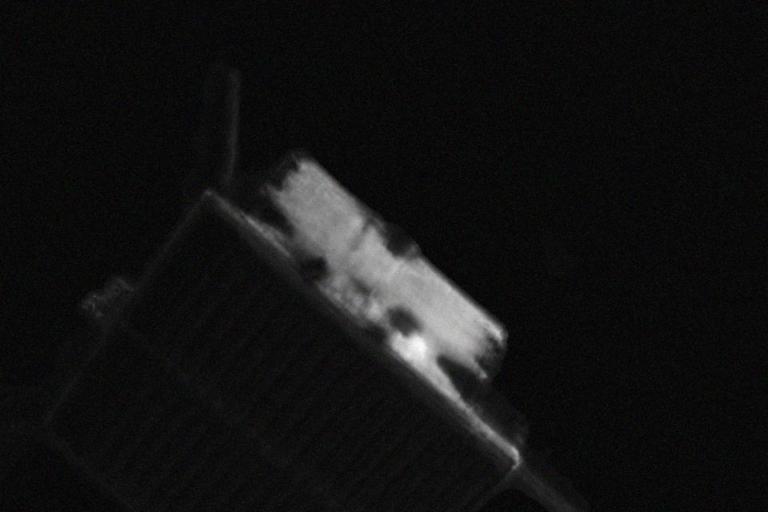}\hfill
    \includegraphics[width=.245\linewidth]{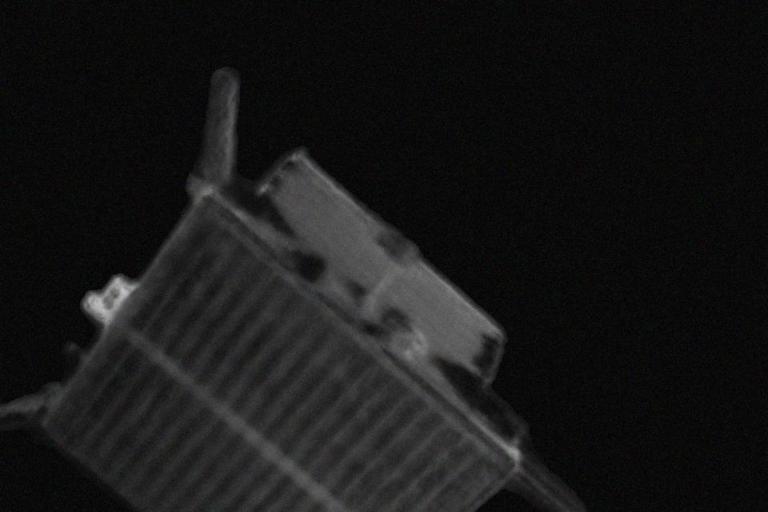}\\
    \caption{Images generated by NeRFs trained on 50, 100, 200 or 500 images (left to right). Even if 50 images are sufficient to recover a fair representation of the target spacecraft, increasing the number of training images reduces the artifacts and enables the recovery of finer details.}
    \label{fig_nerf_number_training_images}
    \vspace{-0.2cm}
\end{figure}

\begin{figure}[t]
    \includegraphics[width=.245\linewidth]{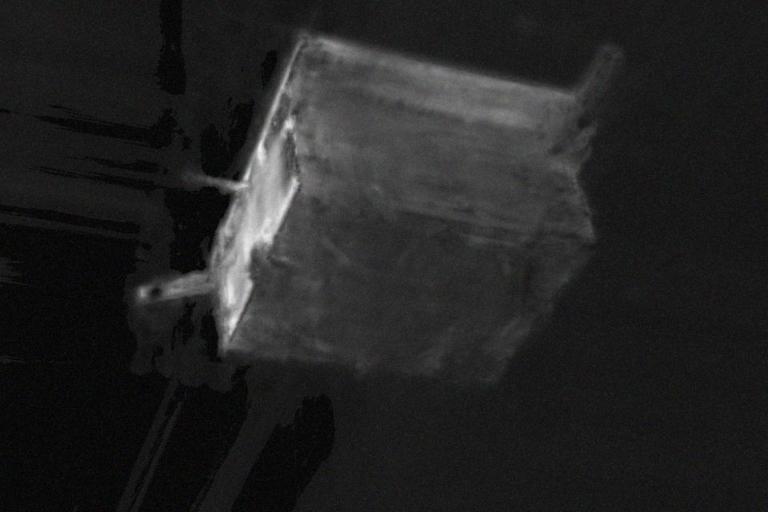}\hfill
    \includegraphics[width=.245\linewidth]{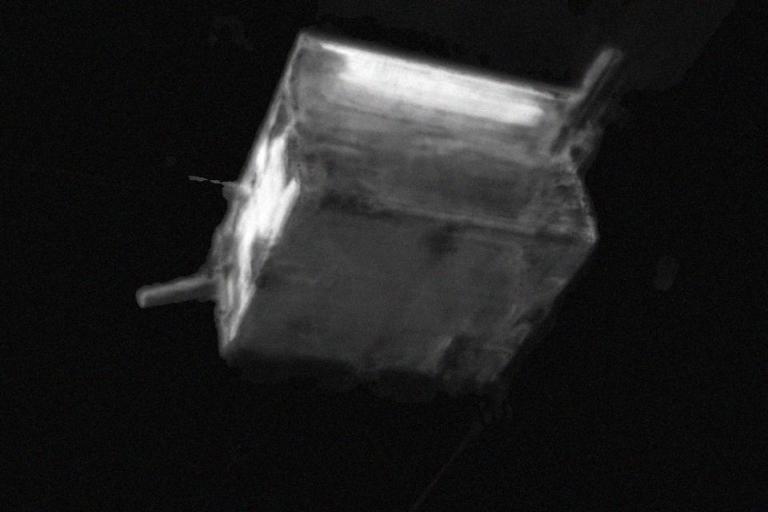}\hfill
    \includegraphics[width=.245\linewidth]{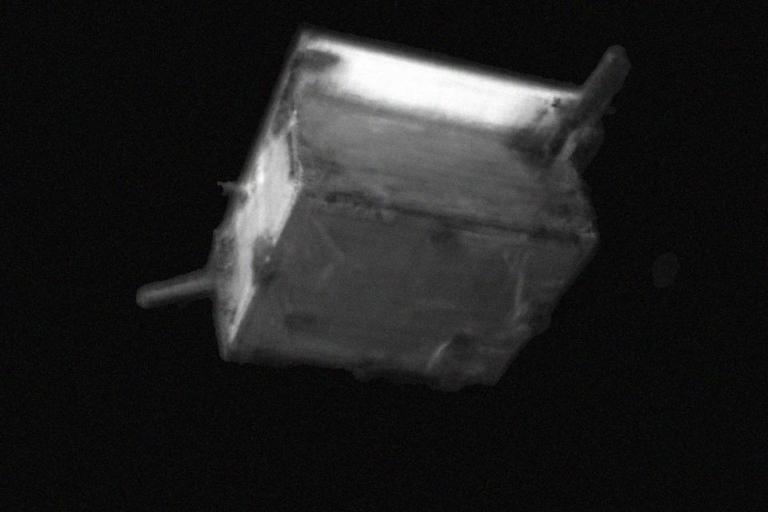}\hfill
    \includegraphics[width=.245\linewidth]{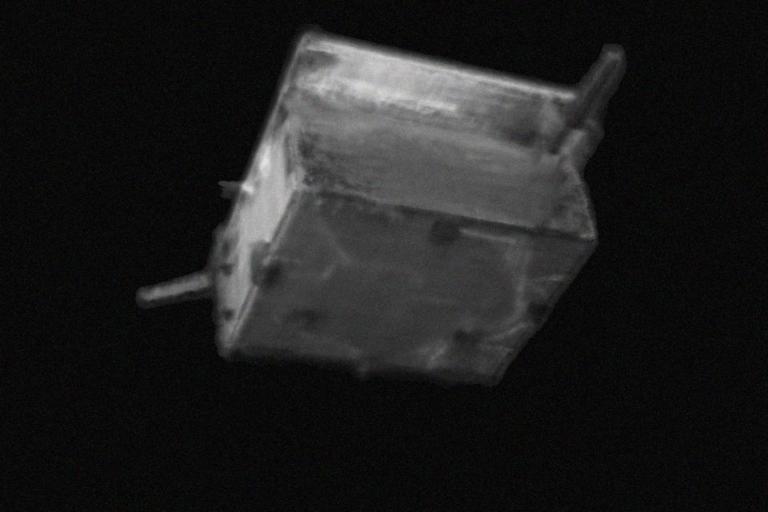}\\    
    \includegraphics[width=.245\linewidth]{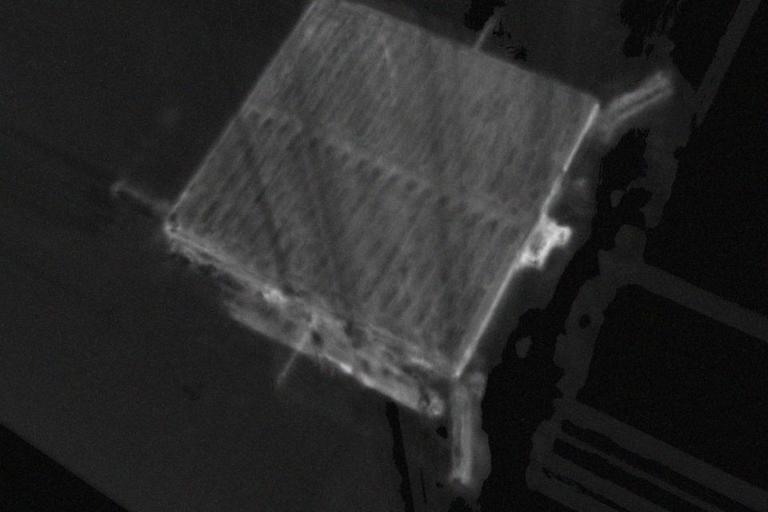}\hfill
    \includegraphics[width=.245\linewidth]{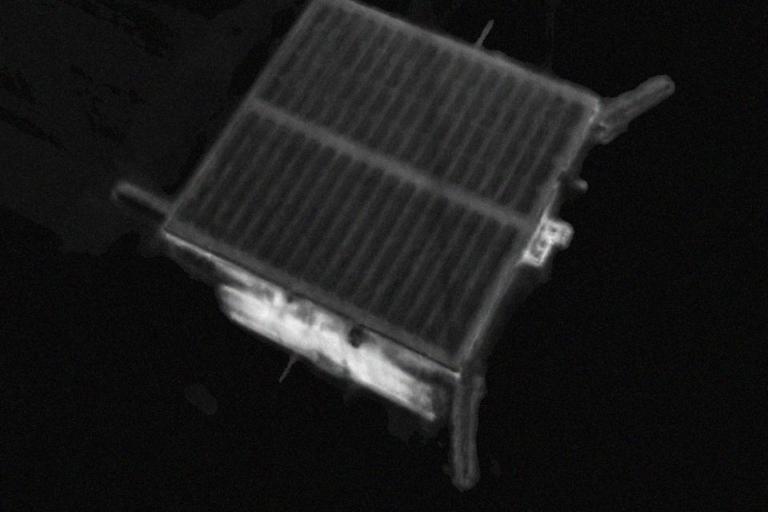}\hfill
    \includegraphics[width=.245\linewidth]{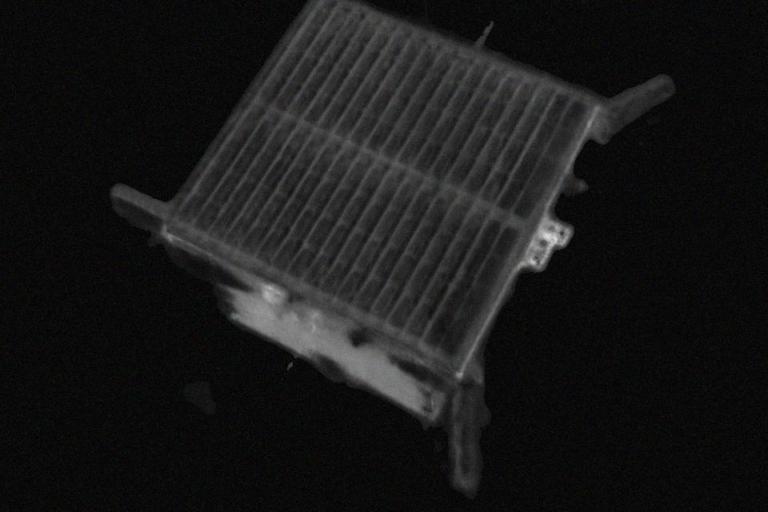}\hfill
    \includegraphics[width=.245\linewidth]{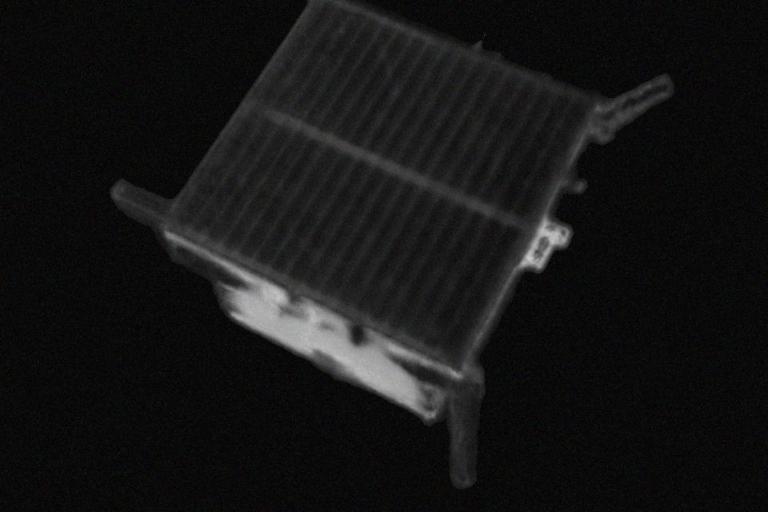}\\    
    \includegraphics[width=.245\linewidth]{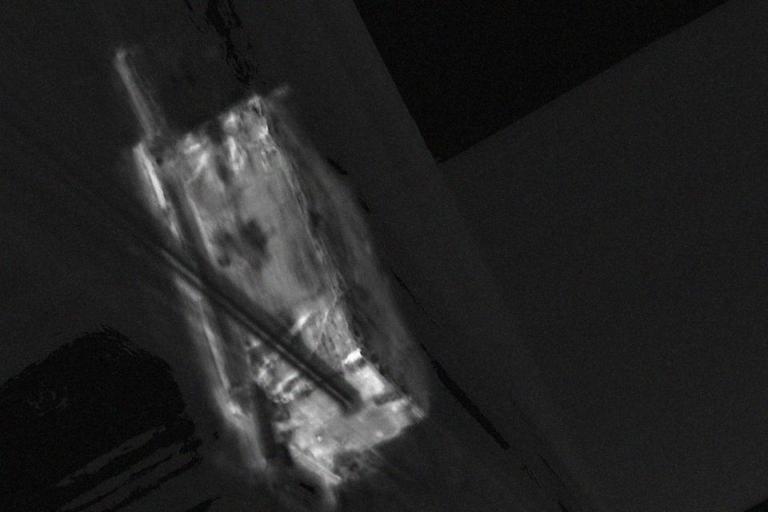}\hfill
    \includegraphics[width=.245\linewidth]{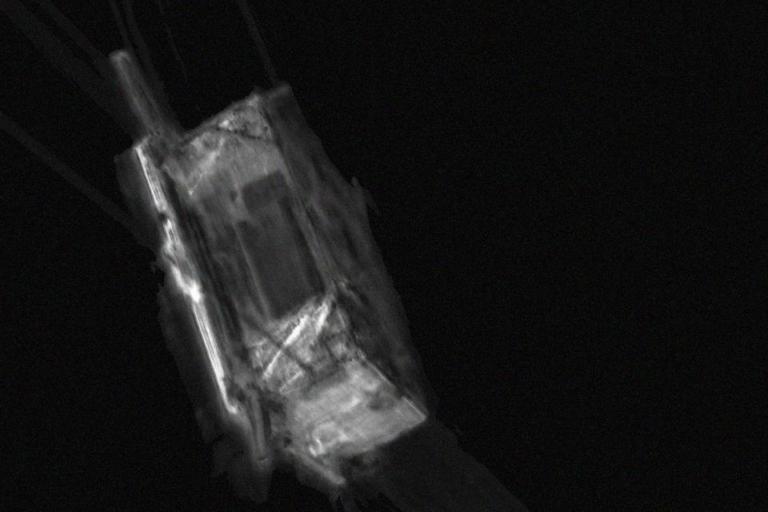}\hfill
    \includegraphics[width=.245\linewidth]{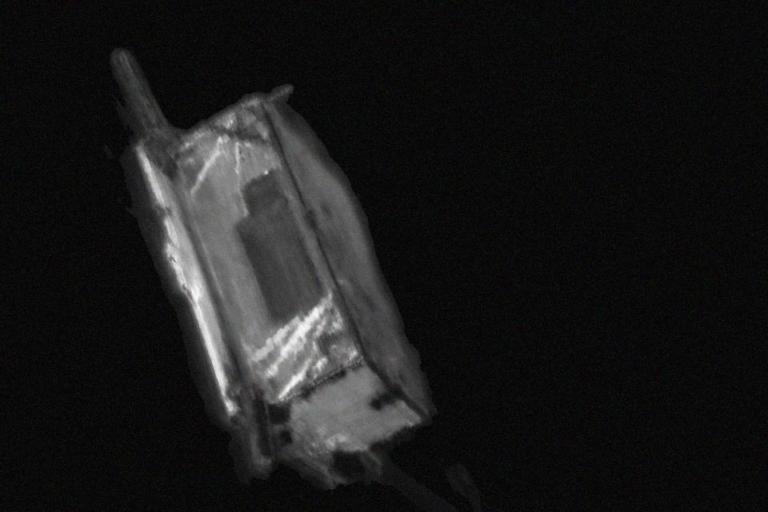}\hfill
    \includegraphics[width=.245\linewidth]{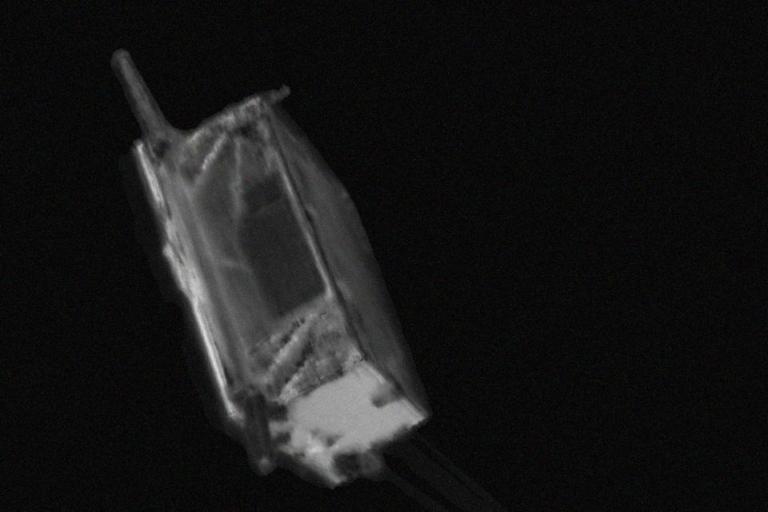}\\    
    \includegraphics[width=.245\linewidth]{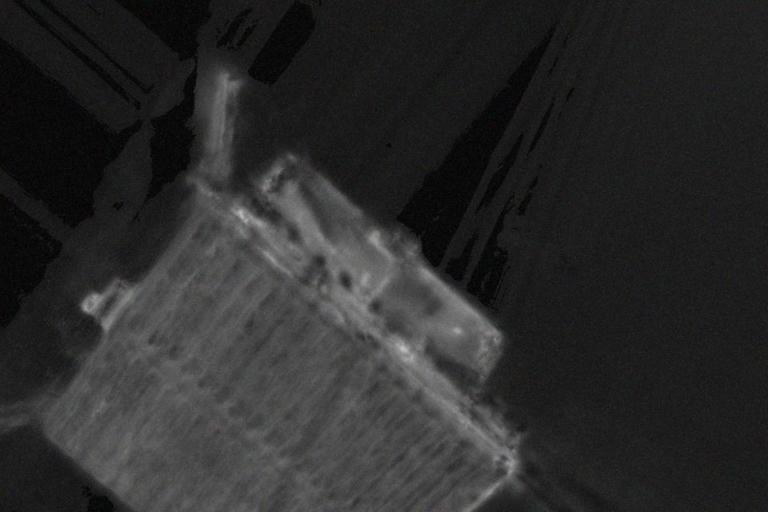}\hfill
    \includegraphics[width=.245\linewidth]{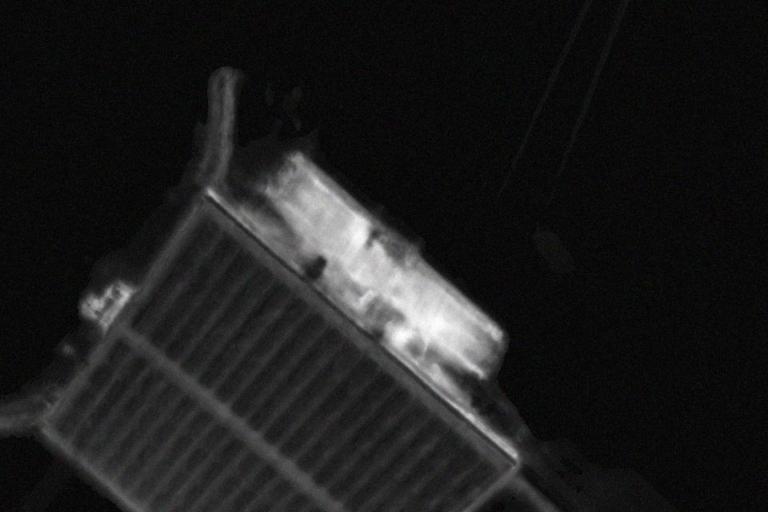}\hfill
    \includegraphics[width=.245\linewidth]{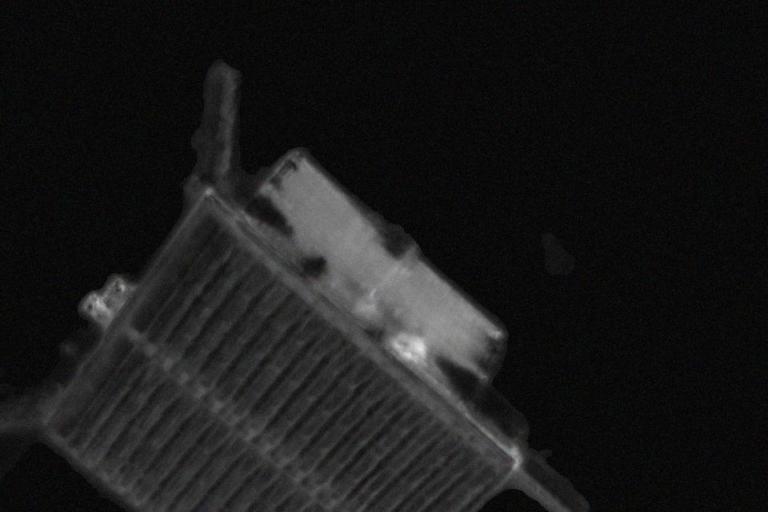}\hfill
    \includegraphics[width=.245\linewidth]{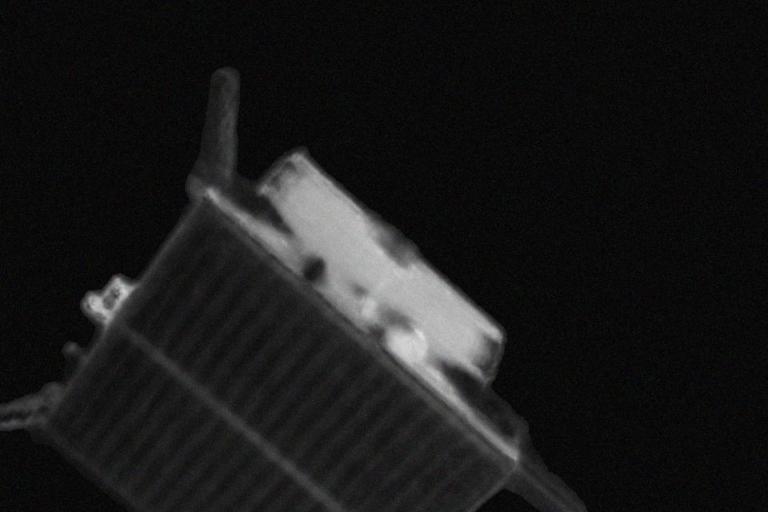}\\
    \caption{Images generated by NeRFs trained on 50, 100, 200 or 500 images (left to right) and rendered using average appearance embeddings. As for the appearance interpolation strategy (see \cref{fig_nerf_number_training_images}), 50 images are sufficient to recover a fair representation of the target spacecraft while finer details can be recovered by increasing the number of training images. Unlike the images generated using randomly interpolated appearance embeddings, the rendered images do not capture the illumination diversity depicted in \Cref{fig_nerf_number_training_images}.}
    \label{fig_nerf_number_training_images_classical}
    \vspace{-0.2cm}
\end{figure}

    \subsection{Evaluation}
    \label{sec_expe_main}
    
        \Cref{fig_graph_number_training_images} compares the SPEED+ score, \ie the sum of the average rotation and normalized translation errors (see \cref{sec_imp_details}), achieved by the SPE network trained either on $N$ spaceborne images directly or on 48,000 images generated by a NeRF trained on those $N$ images. While the baseline, straightforward, approach fails in estimating the pose of the target spacecraft, our approach successfully predicts the pose from a limited set of spaceborne images regardless of the number of images used to train the NeRF. Indeed, as highlighted in \cref{tab_main_results}, while the baseline approach exhibits errors of 99.4\degree-1.20m on \textit{sunlamp} and 95.0\degree-1.31m on \textit{lightbox}, our approach achieves errors of 16.9\degree-0.29m and 7.2\degree-0.20m on those sets. \Cref{fig_graph_number_training_images} also depicts the SPEED+ score obtained with the same SPE network but trained on the 48,000 synthetic images of SPEED+, rendered using the CAD model of the target. Our approach, which is model-agnostic, trained with 500 real images, performs as well as, or even better than, the CAD-dependent one, which achieves errors of 15.7\degree-0.30m and 9.5\degree-0.25m on those sets. Examples of images generated with our method are depicted in \Cref{fig_nerf_number_training_images}.

    \subsection{Ablation Study: Appearance Interpolation Strategy}
    \label{sec_expe_ablation}
        This section explores the impact of the appearance interpolation strategy on the accuracy of the SPE network. For this purpose, \Cref{fig_graph_in_the_wild} depicts the relationship between the number of images used to train the NeRF and the SPEED+ score of the SPE trained on images rendered using either randomly interpolated appearance embeddings or a single appearance embedding that corresponds to the average of the learned appearance embeddings. \Cref{fig_nerf_number_training_images_classical} depicts images rendered by NeRFs using average appearance embeddings. Even if the NeRF recovers the shape of the spacecraft, it can not render the diversity of the illumination conditions encountered on the train set. On both \textit{sunlamp} and \textit{lightbox*}, the SPE trained using interpolated appearance embeddings outperforms the one trained with the average appearance.

\section{Conclusions}
    This paper addressed the problem of estimating the pose, \ie position and orientation, of an unknown target spacecraft relative to a chaser, in the context of autonomous Rendezvous and Proximity Operations. For this purpose, we introduced a method that enables the training of an off-the-shelf spacecraft pose estimation network from a limited set of real images depicting the target. To this end, our method resorts to an in-the-wild NeRF, \ie a Neural Radiance Field that uses learnable appearance embeddings to capture the varying appearance conditions encountered on real images, which is trained on this sparse set of images. It is then used to generate a large training set which is finally used to train the pose estimation network.

    The method was validated on the Hardware-In-the-Loop images of SPEED+~\cite{park2022speed+}. We showed that a pose estimation network trained with our approach does not only perform much better than the baseline solution, which consists in training the network directly on the real images, but also achieves a similar accuracy as the same network trained on images generated using the CAD model of the target. In addition, we highlighted the role of the in-the-wild abilities of the Neural Radiance Fields in the efficiency of the proposed method.


\footnotesize
\section*{Acknowledgements}
The research was funded by Aerospacelab and the Walloon Region through the Win4Doc program. Christophe De Vleeschouwer is a Research Director of the Fonds de la Recherche Scientifique - FNRS. Computational resources have been provided by the Consortium des Équipements de Calcul Intensif (CÉCI), funded by the Fonds de la Recherche Scientifique de Belgique (F.R.S.-FNRS) under Grant No. 2.5020.11 and by the Walloon Region.

\normalsize
\bibliography{references}

\begin{thebibliography}{47}
\providecommand{\natexlab}[1]{#1}
\providecommand{\url}[1]{\texttt{#1}}
\expandafter\ifx\csname urlstyle\endcsname\relax
  \providecommand{\doi}[1]{doi: #1}\else
  \providecommand{\doi}{doi: \begingroup \urlstyle{rm}\Url}\fi

\bibitem[Park et~al.(2022)Park, M{\"a}rtens, Lecuyer, Izzo, and D'Amico]{park2022speed+}
Tae~Ha Park, Marcus M{\"a}rtens, Gurvan Lecuyer, Dario Izzo, and Simone D'Amico.
\newblock Speed+: Next-generation dataset for spacecraft pose estimation across domain gap.
\newblock In \emph{2022 IEEE Aerospace Conference (AERO)}, pages 1--15. IEEE, 2022.

\bibitem[Rossi et~al.(1998)Rossi, Anselmo, Cordelli, Farinella, and Pardini]{rossi1998modelling}
A~Rossi, L~Anselmo, A~Cordelli, P~Farinella, and C~Pardini.
\newblock Modelling the evolution of the space debris population.
\newblock \emph{Planetary and Space Science}, 46\penalty0 (11-12):\penalty0 1583--1596, 1998.

\bibitem[Kessler et~al.(2010)Kessler, Johnson, Liou, and Matney]{kessler2010kessler}
Donald~J Kessler, Nicholas~L Johnson, JC~Liou, and Mark Matney.
\newblock The kessler syndrome: implications to future space operations.
\newblock \emph{Advances in the Astronautical Sciences}, 137\penalty0 (8):\penalty0 2010, 2010.

\bibitem[Forshaw et~al.(2020)Forshaw, Aglietti, Fellowes, Salmon, Retat, Hall, Chabot, Pisseloup, Tye, Bernal, et~al.]{forshaw2020active}
Jason~L Forshaw, Guglielmo~S Aglietti, Simon Fellowes, Thierry Salmon, Ingo Retat, Alexander Hall, Thomas Chabot, Aur{\'e}lien Pisseloup, Daniel Tye, Cesar Bernal, et~al.
\newblock The active space debris removal mission removedebris. part 1: From concept to launch.
\newblock \emph{Acta Astronautica}, 168:\penalty0 293--309, 2020.

\bibitem[Aglietti et~al.(2020)Aglietti, Taylor, Fellowes, Salmon, Retat, Hall, Chabot, Pisseloup, Cox, Mafficini, et~al.]{aglietti2020active}
Guglielmo~S Aglietti, Ben Taylor, Simon Fellowes, Thierry Salmon, Ingo Retat, Alexander Hall, Thomas Chabot, Aur{\'e}lien Pisseloup, Christopher Cox, A~Mafficini, et~al.
\newblock The active space debris removal mission removedebris. part 2: In orbit operations.
\newblock \emph{Acta Astronautica}, 168:\penalty0 310--322, 2020.

\bibitem[Poozhiyil et~al.(2023)Poozhiyil, Nair, Rai, Hall, Meringolo, Shilton, Kay, Forte, Sweeting, Antoniou, et~al.]{poozhiyil2023active}
Mithun Poozhiyil, Manu~H Nair, Mini~C Rai, Alexander Hall, Connor Meringolo, Mark Shilton, Steven Kay, Danilo Forte, Martin Sweeting, Nikki Antoniou, et~al.
\newblock Active debris removal: A review and case study on leopard phase 0-a mission.
\newblock \emph{Advances in Space Research}, 2023.

\bibitem[D’Amico et~al.(2014)D’Amico, Benn, and J{\o}rgensen]{d2014pose}
Simone D’Amico, Mathias Benn, and John~L J{\o}rgensen.
\newblock Pose estimation of an uncooperative spacecraft from actual space imagery.
\newblock \emph{International Journal of Space Science and Engineering 5}, 2\penalty0 (2):\penalty0 171--189, 2014.

\bibitem[Pauly et~al.(2023)Pauly, Rharbaoui, Shneider, Rathinam, Gaudilli{\`e}re, and Aouada]{pauly2023survey}
Leo Pauly, Wassim Rharbaoui, Carl Shneider, Arunkumar Rathinam, Vincent Gaudilli{\`e}re, and Djamila Aouada.
\newblock A survey on deep learning-based monocular spacecraft pose estimation: Current state, limitations and prospects.
\newblock \emph{Acta Astronautica}, 2023.

\bibitem[Park and D’Amico(2023)]{park2023robust}
Tae~Ha Park and Simone D’Amico.
\newblock Robust multi-task learning and online refinement for spacecraft pose estimation across domain gap.
\newblock \emph{Advances in Space Research}, 2023.

\bibitem[Chen et~al.(2019)Chen, Cao, Parra, and Chin]{chen2019satellite}
Bo~Chen, Jiewei Cao, Alvaro Parra, and Tat-Jun Chin.
\newblock Satellite pose estimation with deep landmark regression and nonlinear pose refinement.
\newblock In \emph{Proceedings of the IEEE/CVF International Conference on Computer Vision Workshops}, pages 0--0, 2019.

\bibitem[Proen{\c{c}}a and Gao(2020)]{proencca2020deep}
Pedro~F Proen{\c{c}}a and Yang Gao.
\newblock Deep learning for spacecraft pose estimation from photorealistic rendering.
\newblock In \emph{2020 IEEE International Conference on Robotics and Automation (ICRA)}, pages 6007--6013. IEEE, 2020.

\bibitem[Mildenhall et~al.(2021)Mildenhall, Srinivasan, Tancik, Barron, Ramamoorthi, and Ng]{mildenhall2021nerf}
Ben Mildenhall, Pratul~P Srinivasan, Matthew Tancik, Jonathan~T Barron, Ravi Ramamoorthi, and Ren Ng.
\newblock Nerf: Representing scenes as neural radiance fields for view synthesis.
\newblock \emph{Communications of the ACM}, 65\penalty0 (1):\penalty0 99--106, 2021.

\bibitem[Fridovich-Keil et~al.(2023)Fridovich-Keil, Meanti, Warburg, Recht, and Kanazawa]{fridovich2023k}
Sara Fridovich-Keil, Giacomo Meanti, Frederik~Rahb{\ae}k Warburg, Benjamin Recht, and Angjoo Kanazawa.
\newblock K-planes: Explicit radiance fields in space, time, and appearance.
\newblock In \emph{Proceedings of the IEEE/CVF Conference on Computer Vision and Pattern Recognition}, pages 12479--12488, 2023.

\bibitem[Martin-Brualla et~al.(2021)Martin-Brualla, Radwan, Sajjadi, Barron, Dosovitskiy, and Duckworth]{martin2021nerf}
Ricardo Martin-Brualla, Noha Radwan, Mehdi~SM Sajjadi, Jonathan~T Barron, Alexey Dosovitskiy, and Daniel Duckworth.
\newblock Nerf in the wild: Neural radiance fields for unconstrained photo collections.
\newblock In \emph{Proceedings of the IEEE/CVF Conference on Computer Vision and Pattern Recognition}, pages 7210--7219, 2021.

\bibitem[Huang et~al.(2021)Huang, Zhang, Cui, and Zhang]{huang2021low}
Yefei Huang, Zexu Zhang, Hutao Cui, and Liang Zhang.
\newblock A low-dimensional binary-based descriptor for unknown satellite relative pose estimation.
\newblock \emph{Acta Astronautica}, 181:\penalty0 427--438, 2021.

\bibitem[Ren et~al.(2020)Ren, Jiang, and Wang]{ren2020pose}
Xiaoyuan Ren, Libing Jiang, and Zhuang Wang.
\newblock Pose estimation of uncooperative unknown space objects from a single image.
\newblock \emph{International Journal of Aerospace Engineering}, 2020:\penalty0 1--9, 2020.

\bibitem[Park and D'Amico(2024)]{park2024rapid}
Tae~Ha Park and Simone D'Amico.
\newblock Rapid abstraction of spacecraft 3d structure from single 2d image.
\newblock In \emph{AIAA SCITECH 2024 Forum}, page 2768, 2024.

\bibitem[Pesce et~al.(2017)Pesce, Lavagna, and Bevilacqua]{pesce2017stereovision}
Vincenzo Pesce, Mich{\`e}le Lavagna, and Riccardo Bevilacqua.
\newblock Stereovision-based pose and inertia estimation of unknown and uncooperative space objects.
\newblock \emph{Advances in Space Research}, 59\penalty0 (1):\penalty0 236--251, 2017.

\bibitem[Feng et~al.(2018)Feng, Zhu, Pan, and Liu]{feng2018pose}
Qian Feng, Zheng~H Zhu, Quan Pan, and Yong Liu.
\newblock Pose and motion estimation of unknown tumbling spacecraft using stereoscopic vision.
\newblock \emph{Advances in Space Research}, 62\penalty0 (2):\penalty0 359--369, 2018.

\bibitem[Guo et~al.(2020)Guo, He, Qi, Wu, Hu, Li, and Zhang]{guo2020real}
Jiawei Guo, Yucheng He, Xiaozhi Qi, Guangxin Wu, Ying Hu, Bing Li, and Jianwei Zhang.
\newblock Real-time measurement and estimation of the 3d geometry and motion parameters for spatially unknown moving targets.
\newblock \emph{Aerospace Science and Technology}, 97:\penalty0 105619, 2020.

\bibitem[Matsuka et~al.(2021)Matsuka, Santamaria-Navarro, Capuano, Harvard, Rahmani, and Chung]{matsuka2021collaborative}
Kai Matsuka, Angel Santamaria-Navarro, Vincenzo Capuano, Alexei Harvard, Amir Rahmani, and Soon-Jo Chung.
\newblock Collaborative pose estimation of an unknown target using multiple spacecraft.
\newblock In \emph{2021 IEEE Aerospace Conference (50100)}, pages 1--11. IEEE, 2021.

\bibitem[Harris et~al.(1988)Harris, Stephens, et~al.]{harris1988combined}
Chris Harris, Mike Stephens, et~al.
\newblock A combined corner and edge detector.
\newblock In \emph{Alvey vision conference}, volume~15, pages 10--5244. Citeseer, 1988.

\bibitem[Lowe(2004)]{lowe2004distinctive}
David~G Lowe.
\newblock Distinctive image features from scale-invariant keypoints.
\newblock \emph{International journal of computer vision}, 60:\penalty0 91--110, 2004.

\bibitem[Rublee et~al.(2011)Rublee, Rabaud, Konolige, and Bradski]{rublee2011orb}
Ethan Rublee, Vincent Rabaud, Kurt Konolige, and Gary Bradski.
\newblock Orb: An efficient alternative to sift or surf.
\newblock In \emph{2011 International conference on computer vision}, pages 2564--2571. Ieee, 2011.

\bibitem[Kisantal et~al.(2020)Kisantal, Sharma, Park, Izzo, M{\"a}rtens, and D’Amico]{kisantal2020satellite}
Mate Kisantal, Sumant Sharma, Tae~Ha Park, Dario Izzo, Marcus M{\"a}rtens, and Simone D’Amico.
\newblock Satellite pose estimation challenge: Dataset, competition design, and results.
\newblock \emph{IEEE Transactions on Aerospace and Electronic Systems}, 56\penalty0 (5):\penalty0 4083--4098, 2020.

\bibitem[Park et~al.(2023)Park, M{\"a}rtens, Jawaid, Wang, Chen, Chin, Izzo, and D’Amico]{park2023satellite}
Tae~Ha Park, Marcus M{\"a}rtens, Mohsi Jawaid, Zi~Wang, Bo~Chen, Tat-Jun Chin, Dario Izzo, and Simone D’Amico.
\newblock Satellite pose estimation competition 2021: Results and analyses.
\newblock \emph{Acta Astronautica}, 204:\penalty0 640--665, 2023.

\bibitem[Sharma et~al.(2018)Sharma, Beierle, and D'Amico]{sharma2018pose}
Sumant Sharma, Connor Beierle, and Simone D'Amico.
\newblock Pose estimation for non-cooperative spacecraft rendezvous using convolutional neural networks.
\newblock In \emph{2018 IEEE Aerospace Conference}, pages 1--12. IEEE, 2018.

\bibitem[Park et~al.(2019)Park, Sharma, and D'Amico]{park2019towards}
Tae~Ha Park, Sumant Sharma, and Simone D'Amico.
\newblock Towards robust learning-based pose estimation of noncooperative spacecraft.
\newblock \emph{arXiv preprint arXiv:1909.00392}, 2019.

\bibitem[Legrand et~al.(2022)Legrand, Detry, and De~Vleeschouwer]{legrand2022end}
Antoine Legrand, Renaud Detry, and Christophe De~Vleeschouwer.
\newblock End-to-end neural estimation of spacecraft pose with intermediate detection of keypoints.
\newblock In \emph{European Conference on Computer Vision}, pages 154--169. Springer, 2022.

\bibitem[Lepetit et~al.(2009)Lepetit, Moreno-Noguer, and Fua]{lepetit2009ep}
Vincent Lepetit, Francesc Moreno-Noguer, and Pascal Fua.
\newblock Ep n p: An accurate o (n) solution to the p n p problem.
\newblock \emph{International journal of computer vision}, 81:\penalty0 155--166, 2009.

\bibitem[M{\"u}ller et~al.(2022)M{\"u}ller, Evans, Schied, and Keller]{muller2022instant}
Thomas M{\"u}ller, Alex Evans, Christoph Schied, and Alexander Keller.
\newblock Instant neural graphics primitives with a multiresolution hash encoding.
\newblock \emph{ACM Transactions on Graphics (ToG)}, 41\penalty0 (4):\penalty0 1--15, 2022.

\bibitem[Adamkiewicz et~al.(2022)Adamkiewicz, Chen, Caccavale, Gardner, Culbertson, Bohg, and Schwager]{adamkiewicz2022vision}
Michal Adamkiewicz, Timothy Chen, Adam Caccavale, Rachel Gardner, Preston Culbertson, Jeannette Bohg, and Mac Schwager.
\newblock Vision-only robot navigation in a neural radiance world.
\newblock \emph{IEEE Robotics and Automation Letters}, 7\penalty0 (2):\penalty0 4606--4613, 2022.

\bibitem[Rosinol et~al.(2023)Rosinol, Leonard, and Carlone]{rosinol2023nerf}
Antoni Rosinol, John~J Leonard, and Luca Carlone.
\newblock Nerf-slam: Real-time dense monocular slam with neural radiance fields.
\newblock In \emph{2023 IEEE/RSJ International Conference on Intelligent Robots and Systems (IROS)}, pages 3437--3444. IEEE, 2023.

\bibitem[Deng et~al.(2022)Deng, He, Ye, Duinkharjav, Chakravarthula, Yang, and Sun]{deng2022fov}
Nianchen Deng, Zhenyi He, Jiannan Ye, Budmonde Duinkharjav, Praneeth Chakravarthula, Xubo Yang, and Qi~Sun.
\newblock Fov-nerf: Foveated neural radiance fields for virtual reality.
\newblock \emph{IEEE Transactions on Visualization and Computer Graphics}, 28\penalty0 (11):\penalty0 3854--3864, 2022.

\bibitem[Nguyen et~al.(2024)Nguyen, Bourki, Macudzinski, Brunel, and Bennamoun]{nguyen2024semantically}
Thang-Anh-Quan Nguyen, Amine Bourki, M{\'a}ty{\'a}s Macudzinski, Anthony Brunel, and Mohammed Bennamoun.
\newblock Semantically-aware neural radiance fields for visual scene understanding: A comprehensive review.
\newblock \emph{arXiv preprint arXiv:2402.11141}, 2024.

\bibitem[Yen-Chen et~al.(2021)Yen-Chen, Florence, Barron, Rodriguez, Isola, and Lin]{yen2021inerf}
Lin Yen-Chen, Pete Florence, Jonathan~T Barron, Alberto Rodriguez, Phillip Isola, and Tsung-Yi Lin.
\newblock inerf: Inverting neural radiance fields for pose estimation.
\newblock In \emph{2021 IEEE/RSJ International Conference on Intelligent Robots and Systems (IROS)}, pages 1323--1330. IEEE, 2021.

\bibitem[Giusti et~al.(2022)Giusti, Garcia, Cozine, Suen, Nguyen, and Alimo]{giusti2022marf}
Lorenzo Giusti, Josue Garcia, Steven Cozine, Darrick Suen, Christina Nguyen, and Ryan Alimo.
\newblock Marf: Representing mars as neural radiance fields.
\newblock In \emph{European Conference on Computer Vision}, pages 53--65. Springer, 2022.

\bibitem[Mergy et~al.(2021)Mergy, Lecuyer, Derksen, and Izzo]{mergy2021vision}
Anne Mergy, Gurvan Lecuyer, Dawa Derksen, and Dario Izzo.
\newblock Vision-based neural scene representations for spacecraft.
\newblock In \emph{Proceedings of the IEEE/CVF Conference on Computer Vision and Pattern Recognition}, pages 2002--2011, 2021.

\bibitem[Caruso et~al.(2023)Caruso, Mahendrakar, Nguyen, White, and Steffen]{caruso20233d}
Basilio Caruso, Trupti Mahendrakar, Van~Minh Nguyen, Ryan~T White, and Todd Steffen.
\newblock 3d reconstruction of non-cooperative resident space objects using instant ngp-accelerated nerf and d-nerf.
\newblock \emph{arXiv preprint arXiv:2301.09060}, 2023.

\bibitem[Heintz and Peck(2023)]{heintz2023spacecraft}
Aneesh~M Heintz and Mason Peck.
\newblock Spacecraft state estimation using neural radiance fields.
\newblock \emph{Journal of Guidance, Control, and Dynamics}, pages 1--14, 2023.

\bibitem[Gill et~al.(2007)Gill, D’Amico, and Montenbruck]{gill2007autonomous}
Eberhard Gill, Simone D’Amico, and Oliver Montenbruck.
\newblock Autonomous formation flying for the prisma mission.
\newblock \emph{Journal of Spacecraft and Rockets}, 44\penalty0 (3):\penalty0 671--681, 2007.

\bibitem[Park et~al.(2021)Park, Bosse, and D'Amico]{park2021robotic}
Tae~Ha Park, Juergen Bosse, and Simone D'Amico.
\newblock Robotic testbed for rendezvous and optical navigation: Multi-source calibration and machine learning use cases.
\newblock \emph{arXiv preprint arXiv:2108.05529}, 2021.

\bibitem[Bukschat and Vetter(2020)]{bukschat2020efficientpose}
Yannick Bukschat and Marcus Vetter.
\newblock Efficientpose: An efficient, accurate and scalable end-to-end 6d multi object pose estimation approach.
\newblock \emph{arXiv preprint arXiv:2011.04307}, 2020.

\bibitem[Tan and Le(2019)]{tan2019efficientnet}
Mingxing Tan and Quoc Le.
\newblock Efficientnet: Rethinking model scaling for convolutional neural networks.
\newblock In \emph{International conference on machine learning}, pages 6105--6114. PMLR, 2019.

\bibitem[Tan et~al.(2020)Tan, Pang, and Le]{tan2020efficientdet}
Mingxing Tan, Ruoming Pang, and Quoc~V Le.
\newblock Efficientdet: Scalable and efficient object detection.
\newblock In \emph{Proceedings of the IEEE/CVF conference on computer vision and pattern recognition}, pages 10781--10790, 2020.

\bibitem[Jackson et~al.(2019)Jackson, Abarghouei, Bonner, Breckon, and Obara]{jackson2019style}
Philip~TG Jackson, Amir~Atapour Abarghouei, Stephen Bonner, Toby~P Breckon, and Boguslaw Obara.
\newblock Style augmentation: data augmentation via style randomization.
\newblock In \emph{CVPR workshops}, volume~6, pages 10--11, 2019.

\bibitem[Ioffe and Szegedy(2015)]{ioffe2015batch}
Sergey Ioffe and Christian Szegedy.
\newblock Batch normalization: Accelerating deep network training by reducing internal covariate shift.
\newblock In \emph{International conference on machine learning}, pages 448--456. pmlr, 2015.

\end{thebibliography}


\end{document}